\def\year{2022}\relax
\documentclass[letterpaper]{article} % DO NOT CHANGE THIS
\usepackage{aaai22}  % DO NOT CHANGE THIS
\usepackage{times}  % DO NOT CHANGE THIS
\usepackage{helvet}  % DO NOT CHANGE THIS
\usepackage{courier}  % DO NOT CHANGE THIS
\usepackage[hyphens]{url}  % DO NOT CHANGE THIS
\usepackage{graphicx} % DO NOT CHANGE THIS
\usepackage{natbib}  % DO NOT CHANGE THIS AND DO NOT ADD ANY OPTIONS TO IT
\usepackage{caption} % DO NOT CHANGE THIS AND DO NOT ADD ANY OPTIONS TO IT

\usepackage{amsmath}
\usepackage{amssymb}
\usepackage{subcaption}
\usepackage{booktabs}
\usepackage{threeparttable}
\usepackage{adjustbox}
\usepackage{makecell}
\usepackage{multirow}
\usepackage{xcolor} 
\usepackage{tablefootnote}

\DeclareCaptionStyle{ruled}{labelfont=normalfont,labelsep=colon,strut=off} % DO NOT CHANGE THIS
\usepackage{algorithm}
\usepackage{algpseudocode}

\usepackage{newfloat}
\usepackage{listings}
\floatstyle{ruled}
\newfloat{listing}{tb}{lst}{}
\floatname{listing}{Listing}
\title{DGSS : Domain Generalized Semantic Segmentation using Iterative Style Mining and Latent Representation Alignment}

\author {
 Pranjay Shyam\textsuperscript{\rm 1}, 
 Antyanta Bangunharcana\textsuperscript{\rm 1}, 
 Kuk-Jin Yoon\textsuperscript{\rm 2}, 
 Kyung-Soo Kim\textsuperscript{\rm 1}
}
\affiliations {
 \textsuperscript{\rm 1} Mechatronics, Systems and Control Lab \quad 
 \textsuperscript{\rm 2} Visual Intelligence Lab \\
 Korea Advanced Institute of Science and Technology (KAIST)\\ Daejeon, Republic of Korea. \\
 \{pranjayshyam, antabangun, kjyoon, kyungsookim\}@kaist.ac.kr
}

\begin{document}

\maketitle

\begin{abstract}
Semantic segmentation algorithms require access to well-annotated datasets captured under diverse illumination conditions to ensure consistent performance. However, poor visibility conditions at varying illumination conditions result in laborious and error-prone labeling. Alternatively, using synthetic samples to train segmentation algorithms has gained interest with the drawback of domain gap that results in sub-optimal performance. While current state-of-the-art (SoTA) have proposed different mechanisms to bridge the domain gap, they still perform poorly in low illumination conditions with an average performance drop of - 10.7 mIOU. In this paper, we focus upon single source domain generalization to overcome the domain gap and propose a two-step framework wherein we first identify an adversarial style that maximizes the domain gap between stylized and source images. Subsequently, these stylized images are used to categorically align features such that features belonging to the same class are clustered together in latent space, irrespective of domain gap. Furthermore, to increase intra-class variance while training, we propose a style mixing mechanism wherein the same objects from different styles are mixed to construct a new training image. This framework allows us to achieve a domain generalized semantic segmentation algorithm with consistent performance without prior information of the target domain while relying on a single source. Based on extensive experiments, we match SoTA performance on SYNTHIA $\to$ Cityscapes, GTAV $\to$ Cityscapes while setting new SoTA on GTAV $\to$ Dark Zurich and GTAV $\to$ Night Driving benchmarks without retraining.
\end{abstract}

\section{Introduction}
Supervised learning of semantic segmentation (SS) algorithms requires access to large-scale and well-annotated datasets captured under diverse illumination and weather variations. However, collecting and labeling such large datasets is time-consuming and error-prone in adverse weather conditions. To reduce this data collection and annotation effort, synthetic datasets generated from game engines (GTAV \cite{richter2016playing}, SYNTHIA \cite{ros2016SYNTHIA}) or from simulators (CARLA \cite{Dosovitskiy17}) are increasingly being used for training. However, domain gaps between synthetic and real images result in performance degradation when training solely on synthetic datasets while evaluating real datasets. 

To alleviate this issue, unsupervised domain adaptation (UDA) or domain generalization (DG) mechanisms are utilized wherein DA minimizes the performance gap using an annotated source domain and unlabelled target domain. In contrast, domain generalization focuses on generalization across multiple unknown target domains using either single or multiple source domains. Despite being practically relevant, DG is particularly challenging since to ensure consistent performance across diverse distributions, knowledge of model bias is necessary to minimize prediction inaccuracies. On the contrary, domain adaptation leverages the availability of the target domain to reduce domain shift by relying either on adversarial learning \cite{vu2019advent}, feature alignment in pixel \cite{chen2019crdoco, choi2019self}, feature \cite{chang2019all, hong2018conditional}, output \cite{tsai2018learning} space or self-training \cite{zhang2019category, li2019bidirectional}. While current methods demonstrate the feasibility of using a synthetic dataset for training SS algorithms via domain adaptation, the sensitivity of these algorithms in low illumination conditions or different domains highlights the drawback of these approaches, i.e., despite domain adaption, a trained SS model might not perform well in night conditions.

\begin{figure}[!t]
\renewcommand{\tabcolsep}{1pt} % adjust horizontal space
\renewcommand{\arraystretch}{1} % adjust vertical space
\centering
\begin{tabular}{cccccc}
% GOPR0356_frame_000404_ref_rgb_anon.png
\multicolumn{3}{c}{\includegraphics[width=0.5\columnwidth, height=2.5cm]{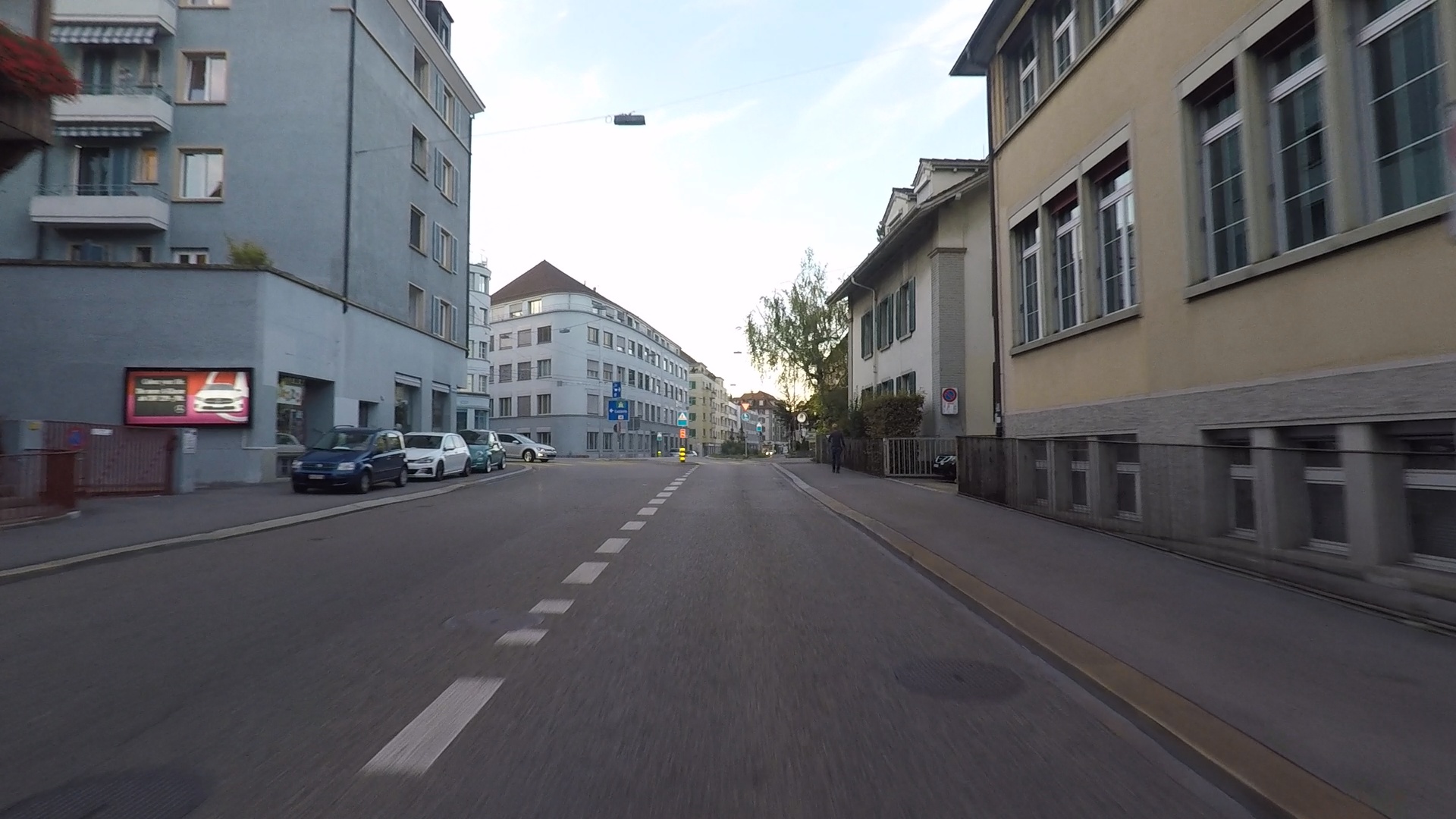}} &
\multicolumn{3}{c}{\includegraphics[width=0.5\columnwidth, height=2.5cm]{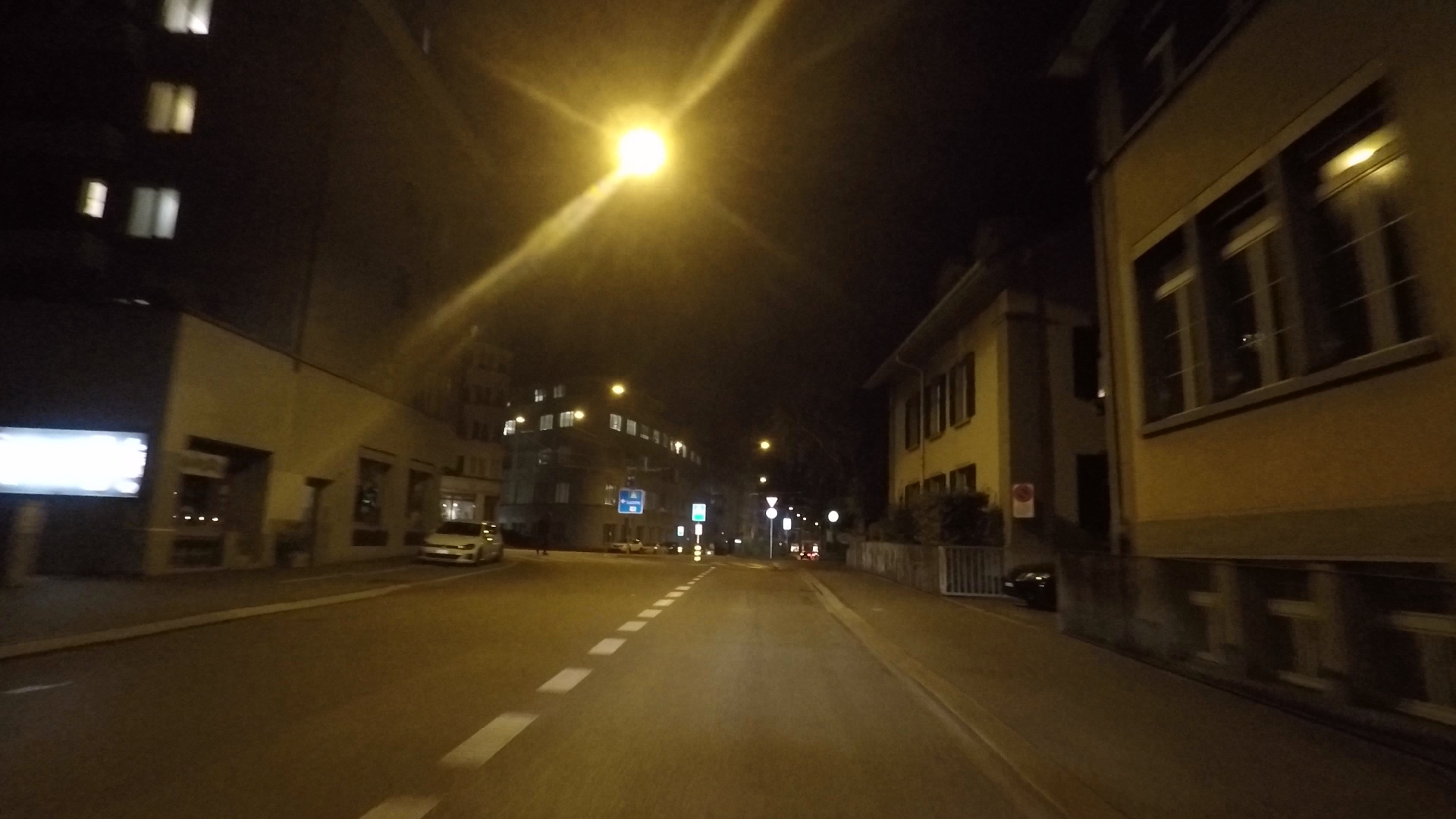}}
\\
\multicolumn{3}{c}{Day} & \multicolumn{3}{c}{Night} \\
\multicolumn{2}{c}{\includegraphics[width=0.33\columnwidth, height=2.5cm]{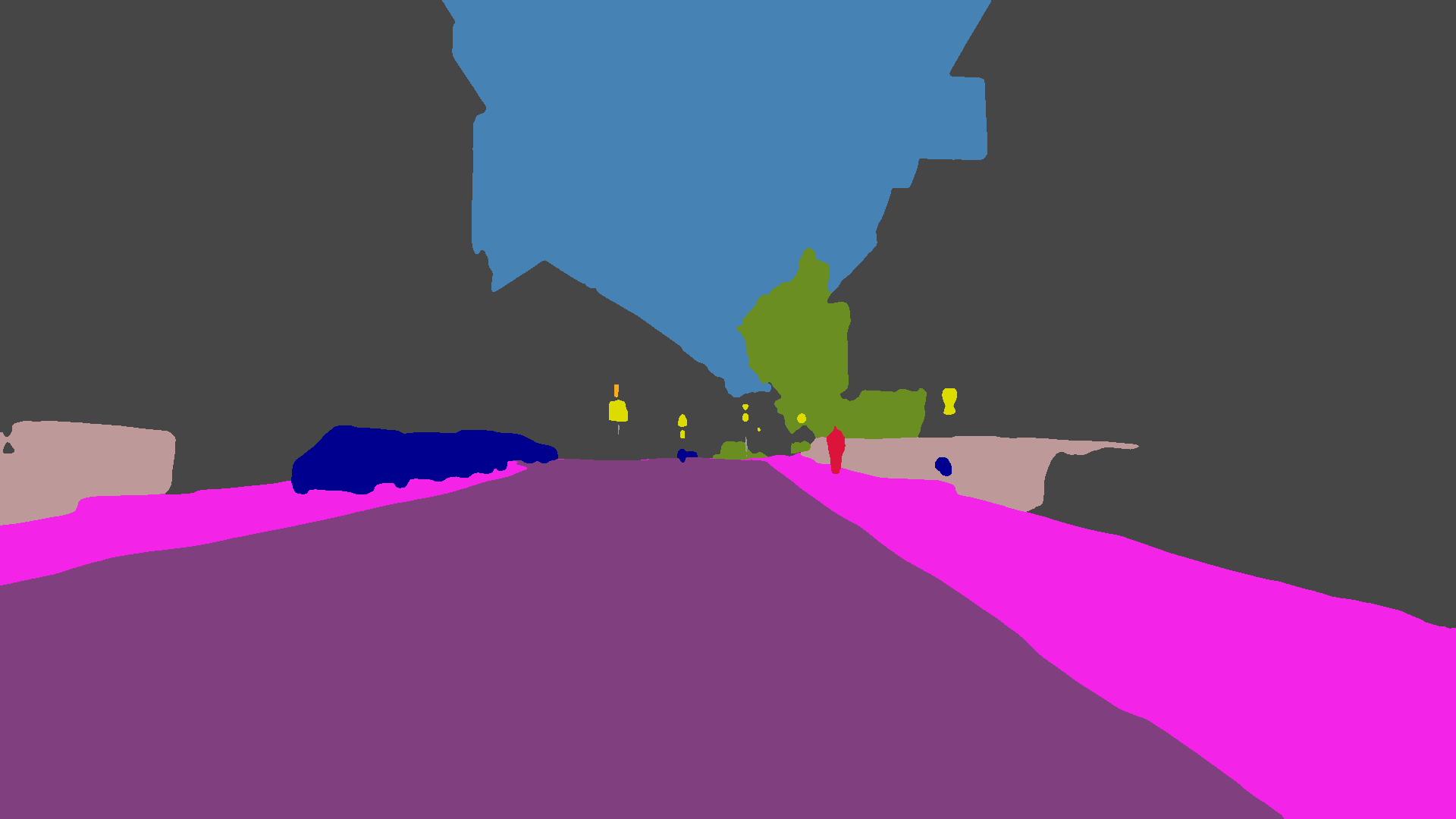}} &
\multicolumn{2}{c}{\includegraphics[width=0.33\columnwidth, height=2.5cm]{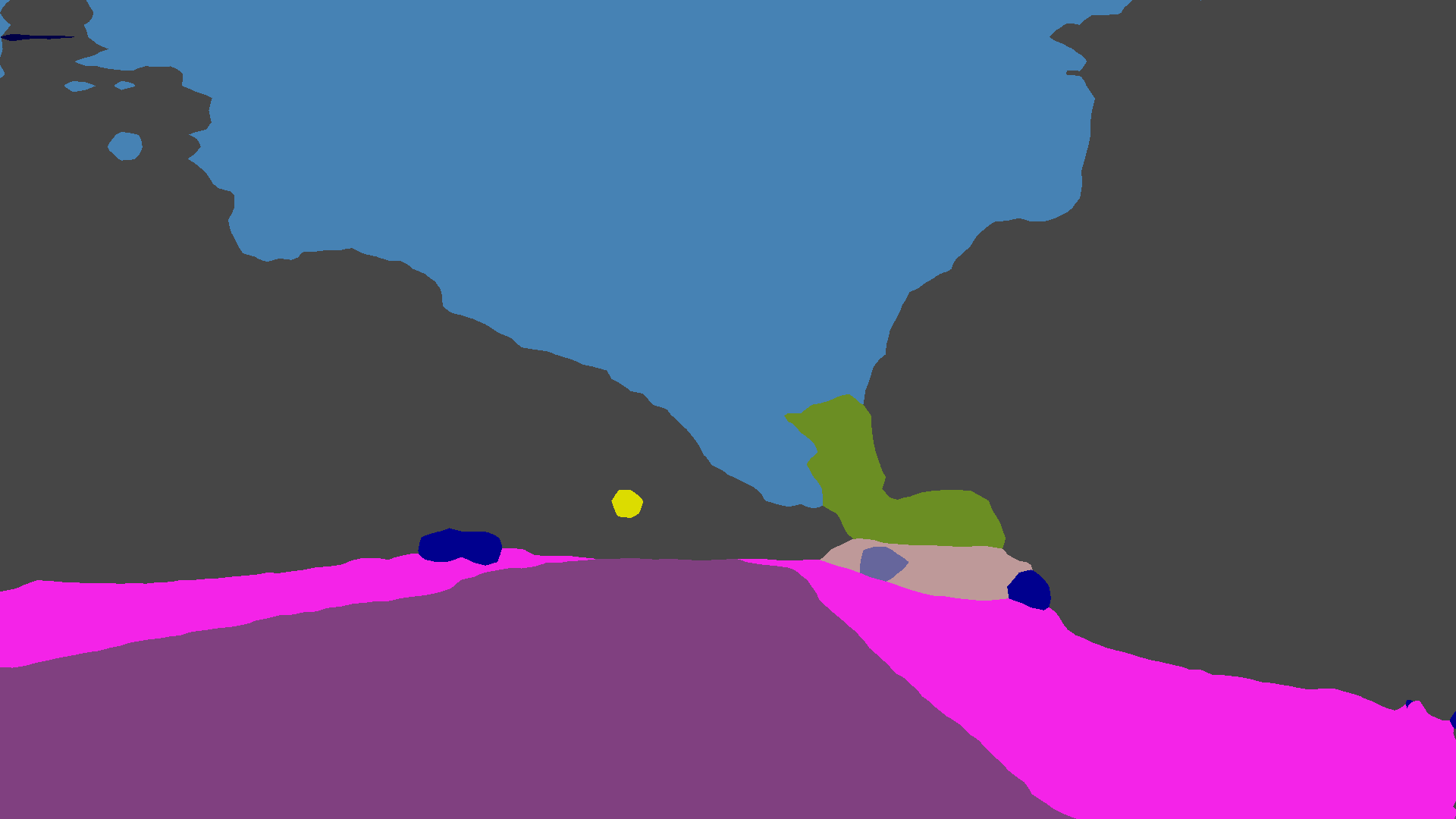}} & 
\multicolumn{2}{c}{\includegraphics[width=0.33\columnwidth, height=2.5cm]{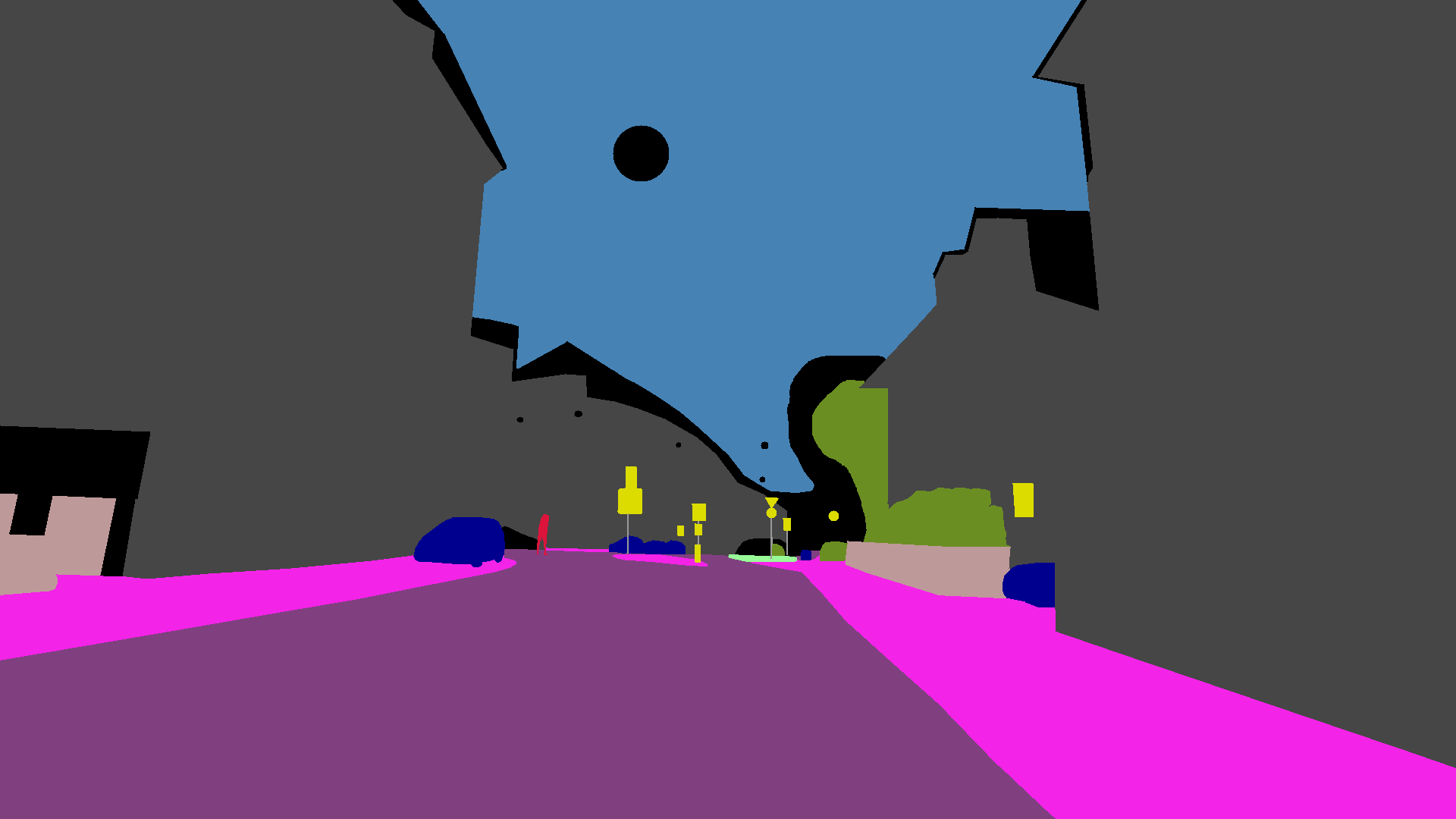}}
\\
\multicolumn{4}{c}{DGSS-Results} & \multicolumn{2}{c}{Night-GT} \\
\end{tabular}
\caption{Results on day and night images from validation subset of Dark-Zurich Dataset.}
\vspace{-5mm}
\label{fig_1}
\end{figure}

In this paper, we propose a target free domain generalization mechanism built upon the notion that features representing the same object category across diverse domains should be similar in latent space irrespective of domain gap. While prior domain generalization works \cite{shyam2021towards, zhao2020maximum, qiao2020learning} relied on leveraging multiple source domains to cover a large feature space, we propose that instead of gathering data from multiple synthetic sources, we could stylize the images using transformations that maximize the distance between the source and stylized image in latent space. This approach allows us to use the same ground truth label and ensures the underlying SS algorithm learns texture invariant structural details. However, using a fixed set of style information results in SS algorithms overfitting these representations and achieves sub-optimal performance. To avoid this, we integrate an iterative style mining approach to identify and use adversarial styles. Upon mining the adversarial styles, we utilize the ground truth labels to align the feature encodings categorically using contrastive learning wherein features belonging to the same category are clustered together and separated from other categories in latent space.  

Inspired by different data augmentation techniques \cite{yun2019cutmix, shyam2021evaluating, ghiasi2021simple}, to maximize intra-class feature distribution, we propose style mixing augmentation wherein given a pool of stylized images we copy-paste different regions to generate an image mosaic such that features belonging to the same category would have different textures. Hence during the optimization process, for the categorical features to be clustered closely, their structural information would be emphasized, resulting in domain invariant characteristics. We demonstrate that the proposed mechanism allows for consistent performance irrespective of domain or illumination changes through extensive experiments. We summarize our contributions as,

\begin{itemize}
 \item We propose an iterative style mining approach to identify adversarial styles to train a semantic segmentation network using a synthetic dataset.
 \item To ensure categorical feature alignment in latent space, we propose a contrastive learning mechanism to introduce push and pull forces to cluster similar features.
 \item For improving structural representation within features, we introduce a style mixing augmentation that results in features from the same category having different styles.
 \item Through extensive experiments on varying domains and illumination conditions, we demonstrate consistent performance using only a single source domain.
\end{itemize}

\section{Related Works}
\noindent
\textbf{Domain Adaptation :} Current approaches for unsupervised domain adaption (UDA) can be formulated either as adversarial learning, image translation or pseudo label based self-training. Adversarial learning-based approaches rely upon a discriminator to align the features either at global \cite{tsai2018learning} or local \cite{vu2019advent} scale by tasking the discriminator to identify the source of the segmented image. Recently \cite{du2019ssf, luo2019taking, wang2020classes} improved this approach by introducing an additional category aware distribution alignment. Another direction for UDA is aligning input images to match target images either using cyclic image translation mechanisms \cite{hoffman2018cycada} or swapping portions of Fourier spectrum of source image with target image \cite{yang2020fda}. Recently pseudo-label based self-training mechanisms have gained increasing interest on their ability to reach higher SoTA on account of generating pseudo labels for target domain and iteratively improving them while improving the performance of SS algorithm. Based on this notion, different algorithms leveraging entropy minimization \cite{chen2019domain, saito2019semi, vu2019advent} was proposed that encourages the SS algorithm to improved predictions on unlabelled data. However, pseudo labels are prone to noise; thus, to reduce the effect of noise \cite{zhang2021prototypical} proposed estimation of pseudo categories and an online correction mechanism to improve the label quality. 

\noindent
\textbf{Illumination Invariance :} While UDA methods have shown promising results, their performance deteriorates significantly when evaluated in low illuminated conditions. Hence \cite{romera2019bridging, sun2019see} proposed a CycleGAN \cite{zhu2017unpaired} based mechanism where cyclic day-to-night mapping is learnt. Subsequently, the training dataset is enlarged to contain images with varying illumination resulting in increased robustness at varying illumination conditions. Recently \cite{wu2021dannet} proposed a relight network to learn a mapping network between different illumination conditions such that low light images are enhanced to improve the performance of the semantic segmentation network. While these approaches result in consistent model performance under varying illumination conditions, a typical synthetic dataset is diverse enough to account for these conditions. Apart from these observations, we concur current SoTA UDA to be sensitive towards illumination variation based on our experiments.

\begin{figure*}[!ht]
\includegraphics[width=0.95\textwidth, height=7.1cm]{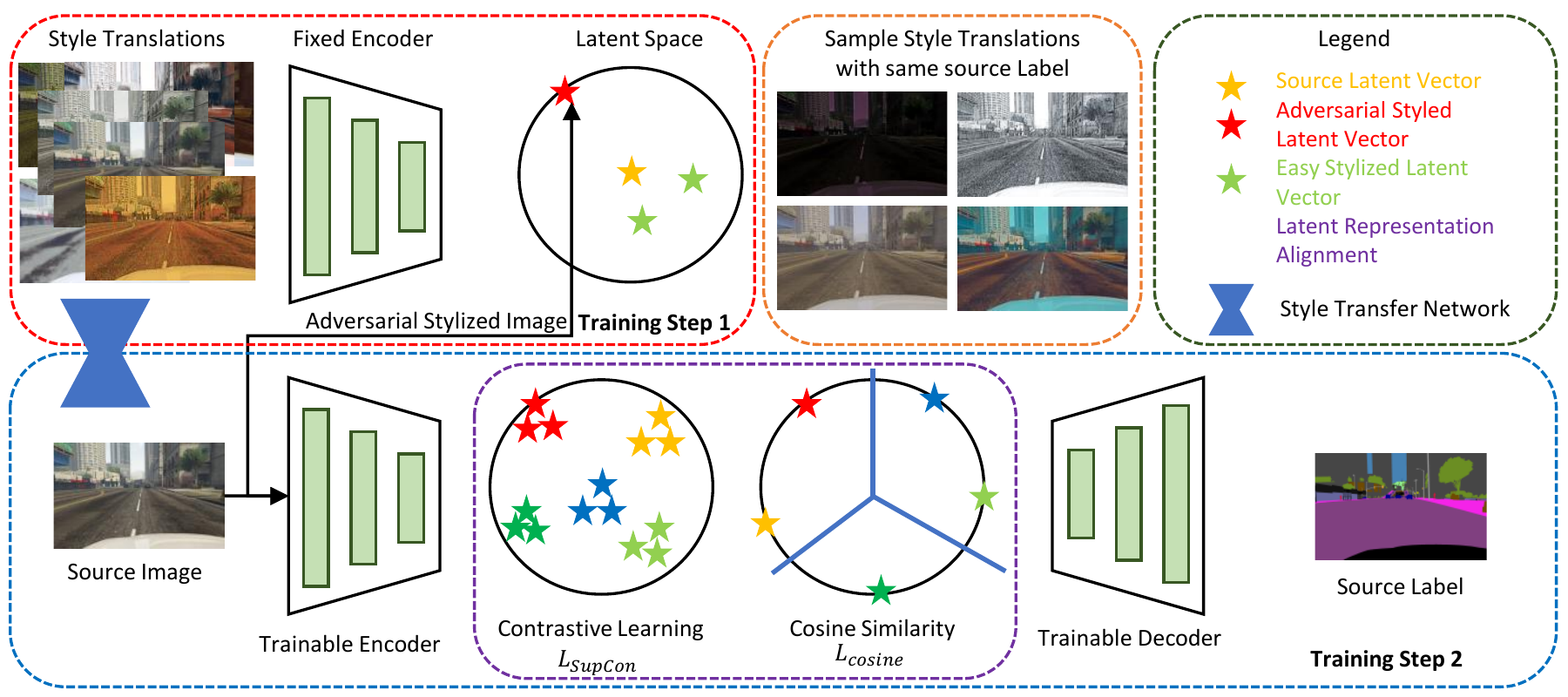}
\caption{Summary of proposed two stage pipeline comprising of iterative style mining and feature representation alignment.}
\vspace{-5mm}
\label{fig_2}
\end{figure*}

\noindent
\textbf{Image stylization for Domain Generalization :} From a practical perspective, it would be beneficial to have a framework that ensures consistent performance irrespective of domain gaps. With this motivation, \cite{muandet2013domain} proposed to learn domain invariant features using a kernel algorithm that minimizes the distributions between multiple distributions. Following this \cite{muandet2013domain} proposed to learn features that are domain invariant by using multiple source domains. Apart from prior approaches that aim to achieve domain generalization by learning domain invariant features, another way to achieve domain generalization is by way of data augmentation. Specifically \cite{jackson2019style} proposed a style augmentation approach to preserve the structure of an image while modifying its texture using neural style transfer resulting in improved robustness of image classification and regression tasks towards domain shift. \cite{chattopadhyay2020learning} extends this approach by learning both domain variant and invariant features in a balanced manner such that the underlying model can perform well throughout. While domain generalization has gained interest for different tasks such as classification, we explore generation of multiple adversarial domains for semantic segmentation simply by the process of image stylization.

\section{Proposed Methodology}
Our framework uses $N$ paired image samples $(x)$ with corresponding categorical labels $(y)$ from single source domain $(S = \{x_i, y_i\}_{i=1}^N)$ to train a semantic segmentation algorithm $F(\theta, C)$ with trainable parameters $\theta$ and maximum categories $C$, such that it retains performance in unseen domains by learning structural representations. For this we follow a two step process wherein we first identify styles that ensure maximum domain gap in feature space. Subsequently we apply these styles on training images and use categorical labels to align similar features in latent space. The complete pipeline is visually summarized in Fig. \ref{fig_2}. 

% \begin{algorithm}
% \caption{DGSS Algorithm}
% \begin{algorithmic}%[1]
% \Require Source Dataset $D(I, N)$ with Images $I$ and Labels $L$
% \Require Style Bank $(SB)$ with $k$ styles
% \Require Style Transfer Network $(STN(SB_k, I)$ accepting a style and image
% \Require ImageNet Pretrained SS Encoder $(S_{Enc}(.))$

% \Comment{Begin Style Mining}
% \State Adversarial Style $ \gets max\{ L1(I, STN(SB_i, I)) \forall i \in k\}$ 

% \Comment{Perform Feature Alignment}
% \State $X \gets x$
% \State $N \gets n$
% \While{$N \neq 0$}
% \If{$N$ is even}
%  \State $X \gets X \times X$
%  \State $N \gets \frac{N}{2}$  
% \ElsIf{$N$ is odd}
%  \State $y \gets y \times X$
%  \State $N \gets N - 1$
% \EndIf
% \EndWhile
% \end{algorithmic}
% \label{algo1}
% \end{algorithm}

\begin{table*}[!th]
\scriptsize
\centering
\begin{adjustbox}{width=\textwidth}
\begin{tabular}{l|l|cccccccccccccccc|c|c}
\Xhline{3\arrayrulewidth} \noalign{\vskip 1pt}
\multirow{5}{*}{\rotatebox{90}{SYNTHIA $\to$ Cityscapes}} & Method  &  
\rotatebox{90}{road} &  
\rotatebox{90}{sidewalk} &  
\rotatebox{90}{building} & 
\rotatebox{90}{wall*} & 
\rotatebox{90}{fence*} & 
\rotatebox{90}{pole*} & 
\rotatebox{90}{light} & 
\rotatebox{90}{sign} & 
\rotatebox{90}{vegetation} & 
\rotatebox{90}{sky} & 
\rotatebox{90}{person} & 
\rotatebox{90}{rider} & 
\rotatebox{90}{car} & 
\rotatebox{90}{bus} & 
\rotatebox{90}{motorbike} & 
\rotatebox{90}{bike} & 
mIoU & 
mIoU* \\
\Xcline{2-20}{2\arrayrulewidth} \noalign{\vskip 1pt}
& Source Only & 55.6 & 23.8 & 74.6 & 9.2 & 0.2 & 24.4 & 6.1 & 12.1 & 74.8 & 79.0 & 55.3 & 19.1 & 39.6 & 23.3 & 13.7 & 25.0 & 33.5 & 38.6 \\
\Xcline{2-20}{2\arrayrulewidth} \noalign{\vskip 1pt}
& IAST & 81.9 & 41.5 & \underline{83.3 }  & 17.7 & \textbf{4.6 } & 32.3 & \underline{30.9 }  & 28.8 & 83.4 & \underline{85.0 }  & 65.5 & \textbf{30.8 } & 86.5 & 38.2 & 33.1 & 52.7 & 49.8 & \underline{57.0 }\\
& MetaCorrection & \textbf{92.6 } & \underline{52.7 }  & 81.3 & 8.9  & \underline{2.4 }  & 28.1 & 13.0 & 7.3  & 83.5 & \underline{85.0 }  & 60.1 & 19.7 & 84.8 & 37.2 & 21.5 & 43.9 & 45.1 & 52.5 \\
& PixMatch  & \underline{92.5 } & \textbf{54.6 } & 79.8 & 4.7  & 0.0 & 24.1 & 22.8 & 17.8 & 79.4 & 76.5 & 60.8 & 24.7 & 85.7 & 33.5 & 26.4 & \textbf{54.4 } & 46.1 & 54.5 \\
& ProDA& 87.8 & 45.7 & \textbf{84.6 } & \textbf{37.1 } & 0.6 & \textbf{44.0 } & \textbf{54.6 } & \textbf{37.0 } & \textbf{88.1 } & 84.4 & \textbf{74.2 } & 24.3 & \textbf{88.2 } & \textbf{51.1 } & \textbf{40.5 } & 45.6 & \textbf{55.5 } & \textbf{62.0 } \\
& FDA  & 84.2 & 35.1 & 78.0 & 6.1  & 0.4 & 27.0 & 8.5  & 22.1 & 77.2 & 79.6 & 55.5 & 19.9 & 74.8 & 24.9 & 14.3 & 40.7 & 40.5 & 48.0 \\
& DGSS & 90.8 & 50.0 & \textbf{84.6}  & \underline{24.8 }  & 1.9 & \underline{38.4 }  & 29.7 & \underline{36.3}& \underline{86.0 }  & \textbf{87.9 } & \underline{67.5 }  & \underline{30.7 }  & 87.3 & 47.3 & 30.2 & \underline{54.0 }  & \underline{52.3}& \underline{\textbf{62.0}}  \\

\Xhline{3\arrayrulewidth} \noalign{\vskip 1pt}
\end{tabular}
\end{adjustbox}
\caption{Performance Comparison with different SoTA domain adaptation algorithms for SYNTHIA $\to$ Cityscapes task. mIoU* denotes the mean IoU of 13 classes, excluding the classes marked by the asterisk.}
\vspace{-5mm}
\label{tab_3}
\end{table*}

\noindent
\textbf{Iterative Style Mining :} Prior domain generalization approaches relied upon data from multiple sources to ensure domain diversity using which domain invariant features are extracted for a given task. However in case of semantic segmentation, access to high quality diverse dataset is restricted. Alternatively, we propose to generate a large synthetic domain by identifying styles that maximize the feature distance between source and stylized images. As the styled features would be far apart from the source features while capturing the same scene, they can be treated as adversarial samples wherein the decoder of the SS algorithm would predict incorrect categories. Hence during training we determine different styles that maximize the distance between features of source and stylized features and use this style for training the underlying SS network without the need to generate a new label. To ensure the underlying SS algorithm learns the structural representation, we would vary the textural content of an image, hence to ensure a wide diversity of textural patterns during style mining process, we use Paintings \cite{4545847} and Textures \cite{cimpoi14describing} as style translation sources. 

\noindent
For our implementation, we train a lightweight style translation algorithm $Sty(x_i, a)$ to obtain a style bank $(a)_{1}^A$ comprising 20 $(=A)$ styles from 10 paintings and 10 textures. Given the style bank, we stylize a mini-batch of 12 training images of size 512 $\times$ 512, and use the frozen encoder of the SS algorithm $(S_{enc}(.))$ to generate feature embeddings corresponding to different stylized images along with source image $(x_i)$. We then compute the $L1$ distance between the stylized and source features to identify adversarial styles that are subsequently used to train the semantic segmentation network. During training it is expected for the underlying SS algorithm to extract features robust towards adversarial styles. Hence style mining is performed iteratively, throughout the training cycle to continuously obtain adversarial styles.

\noindent 
To ensure the style mining operation doesn't create a bottleneck during training, we pre-train a Transformer ResNext Network, Pruned by a factor of 1.0 from \cite{rusty2018faststyletransfer} for 25 randomly chosen textures and paintings with each model containing 63,459 parameters. Furthermore when inferencing these models we use fp16 to further increase processing speed. 

\noindent
\textbf{Latent Representation Alignment :} Irrespective of domain variations, features representing same categories should be clustered closely and separated from other clusters. The ability of the SS encoder to represent features resulting in such a segregation would unequivocally boost the segmentation quality. Hence to enforce such a segregation, we introduce latent representation alignment using prior categorical information. While earlier works \cite{toldo2020unsupervised} have proposed a similar mechanism, they relied on prior source and target information resulting in performance retention only on known target domain whereas our method results in consistent performance across unknown domains. To enforce this segregation we utilize supervised constrastive learning \cite{khosla2020supervised} (SupCon) wherein the image is first converted into feature matrix using the SS encoder $(S_{enc}(.)$ from which pixel-level category based features $(S_{enc}(x) \to z)$ are extracted using corresponding label. Hence for an anchor pixel $(z_a)$ from pixel set $I$, based on category, features are divided into m-positive $(z_p^m)$ and n-negatives pixels $(z_n^n)$. Thus the loss is formulated as,

\vspace{-4mm}
\begin{equation}
\begin{split}
\centering
 \mathrm{L}_{SupCon} = & \\
 \sum_{i \in I} -log\Bigg\{ &\frac{1}{|M(i)|}\sum_{p \in M(i)} \frac{exp(z_i*z_p/\tau)}{\sum_{a \in N(i)} exp(z_i*z_a/\tau)}  \Bigg\}
\end{split}
\end{equation}

% \noindent
% Computing loss on encoded features with a lower spatial size compared to input images, allows contrastive learning to work without any memory overflow. 

\noindent 
While we ensure clustering of similar features to ensure they are far apart in feature space, we utilize cosine similarity to enforce perpendicularity between different features such that distance between features of different classes is maximum. Hence for each unique category cluster we compute the centroid after L1 normalization of features and subsequently use centroids $(c_j \in [1, C])$ of different categories to compute cosine similarity with each feature vector following,

\vspace{-4mm}
\begin{gather}
\mathrm{L_{cosine}} = \sum_{j=1}^C \sum_{i=1}^{m*n} \frac{c_j * z_i}{\parallel c_j \parallel _2 \cdot \parallel z_i \parallel_2}
\end{gather}

\noindent
This formulation allows in ensuring high distance between different category aware features and category centroids. The combination of supervised contrastive learning and cosine similarity allows in extracting features that exhibit the characteristic wherein same category features are close and far away from different category features.

\noindent
\textbf{Style Mixing :} While we use image stylization as an alternative for multiple annotated sources that is necessary to ensure performance generalization. The diversity of objects present within the training batch could be small due to class imbalance. To overcome this propose style mixing augmentation that performs cut-mix \cite{yun2019cutmix} and copy-paste \cite{ghiasi2021simple} operations. Specifically while cut-mix operation focuses on extracting regions from an image and pasting onto another, copy-paste  performs instance aware superimposition. Thus while cut-mix operation acts as a strong regularizer by creating inconsistent textural variations within the image, copy-paste mechanism results in images wherein different instances of same categorical features have different textural properties. Hence using these two augmentations jointly would ensure strong regularization effect while boosting the presence of structural features within the latent representation. We provide qualitative results of style mixing in supplementary.

\noindent
\textbf{Complete Training Objective :} While our approach is generalizable to any semantic segmentation archtiecture, for our analysis following prior works we utilize DeepLabv2 archtiecture with ResNet-101 and VGG-16 as backbones. We implement the proposed framework in PyTorch and initialize the underlying SS encoder with ImageNet \cite{5206848} and train the framework on system with a single Titan-V GPU (12GB) and 32GB RAM. We optimize the framework using Adam \cite{kingma2014adam} with an initial learning rate of 0.0001 and batch size of 1 with GTA-V input resized to 1280 $\times$ 720 owing to GPU memory constraints. Furthermore for each source sample we perform style mining once for epoch and utilize the stylized image with input image, resulting in an effective batch size of 2. We additionally use augmentations such as random flipping and color jittering as augmentations to avoid over fitting and train the models for 80k iterations following the loss objective,

\vspace{-4mm}
\begin{gather}
\mathrm{L} = \lambda_1*\mathrm{L}_{SupCon} + \lambda_2*\mathrm{L}_{cosine}
\end{gather}

\noindent 
here $\lambda_1$ and $\lambda_2$ are weight balances for loss functions which are set to 1 following ablation studies.

\begin{table*}[!th]
\scriptsize
\centering
\begin{adjustbox}{width=\textwidth}
\begin{tabular}{l|l|ccccccccccccccccccc|c}
\Xhline{3\arrayrulewidth} \noalign{\vskip 1pt}
\multirow{5}{*}{\rotatebox{90}{GTAV $\to$ Cityscapes}} & Method &  
\rotatebox{90}{road} &  
\rotatebox{90}{sidewalk} &  
\rotatebox{90}{building} & 
\rotatebox{90}{wall} & 
\rotatebox{90}{fence} & 
\rotatebox{90}{pole} & 
\rotatebox{90}{light} & 
\rotatebox{90}{sign} & 
\rotatebox{90}{vegetation} & 
\rotatebox{90}{terrain} & 
\rotatebox{90}{sky} & 
\rotatebox{90}{person} & 
\rotatebox{90}{rider} & 
\rotatebox{90}{car} & 
\rotatebox{90}{truck} & 
\rotatebox{90}{bus} & 
\rotatebox{90}{train} & 
\rotatebox{90}{motorbike} & 
\rotatebox{90}{bike} & 
mIoU \\
\Xcline{2-22}{2\arrayrulewidth} \noalign{\vskip 1pt}
& Source Only & 75.8 & 16.8 & 77.2 & 12.5 & 21.0 & 25.5 & 30.1 & 20.1 & 81.3 & 24.6 & 70.3 & 53.8 & 26.4 & 49.9 & 17.2 & 25.9 & 6.5 & 25.3 & 36.0 & 36.6 \\
\Xcline{2-22}{2\arrayrulewidth} \noalign{\vskip 1pt}
& IAST& \textbf{94.1} & \textbf{58.8} & 85.4 & \underline{39.7}  & 29.2& 25.1& 43.1& 34.2& 84.8& 34.6& \textbf{88.7} & 62.7& 30.3& 87.6& 42.3& 50.3& \textbf{24.7} & 35.2& 40.2& 52.2 \\
& MetaCorrection & 92.8& \textit{58.1} & \underline{86.2}  & \underline{39.7}  & 33.1& 36.3& 42.0& 38.6& 85.5& 37.8& 87.6& 62.8& 31.7& 84.8& 35.7& 50.3& 2.0 & 36.8& 48.0& 52.1 \\
& PixMatch & 81.0& 33.4& 84.3& 32.9& 27.6& 25.7& 38.3& \underline{47.0}  & 86.5& 36.9& 84.9& 64.6& 28.7 & 5.8 & 42.3& 40.2& 1.5 & 33.7& 41.8& 48.3 \\
& ProDA & 87.8& 56.0& 79.7& \textbf{46.3} & \textbf{44.8} & \textbf{45.6} & \textbf{53.5} & \textbf{53.5} & \textbf{88.6} & \textbf{45.2} & 82.1& \textbf{70.7} & \textbf{39.2} & \underline{88.8}  & \textbf{45.5} & \textbf{59.4} & 1.0 & \textbf{48.9} & \textbf{56.4} & \textbf{57.5}  \\
& FDA & 92.5& 53.3& 82.4& 26.5& 27.6& 36.4& 40.6& 38.9& 82.3& 39.8& 78.0& 62.6& 34.4& 84.9& 34.1& 53.1& \underline{16.9}  & 27.7& 46.4& 50.4 \\
& DGSS& \underline{93.0}  & 52.5& \textbf{86.5} & 35.3& \underline{38.5}  & \underline{40.3}  & \underline{44.7}  & 33.9& \underline{86.9}  & \underline{43.9}  & \underline{87.9}& \underline{67.9}  & \underline{37.8}  & \textbf{89.0} & \underline{44.3}& 52.8& 1.4 & \underline{42.2}  & \underline{55.5}  & \underline{54.4} \\
\Xhline{3\arrayrulewidth} \noalign{\vskip 1pt}
\end{tabular}
\end{adjustbox}
\caption{Performance Comparison with different SoTA domain adaptation algorithms for GTAV $\to$ Cityscapes scenario.}
\label{tab_2}
\end{table*}

\begin{figure*}[!th]
\renewcommand{\tabcolsep}{1pt} % adjust horizontal space
\renewcommand{\arraystretch}{1} % adjust vertical space
\scriptsize
\centering
\begin{adjustbox}{width=\textwidth}

\begin{tabular}{ccccc}
\multicolumn{5}{c}{\includegraphics[width=\textwidth, height=1cm]{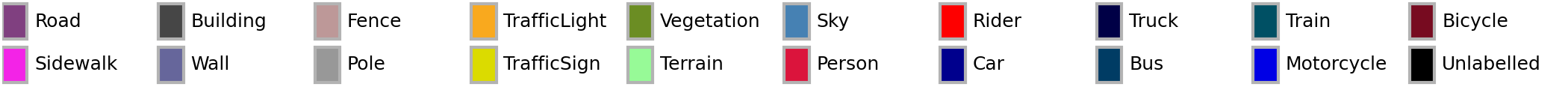}} \\
% lindau_000000_000019_leftImg8bit.png
\includegraphics[width=0.2\textwidth, height=1.75cm]{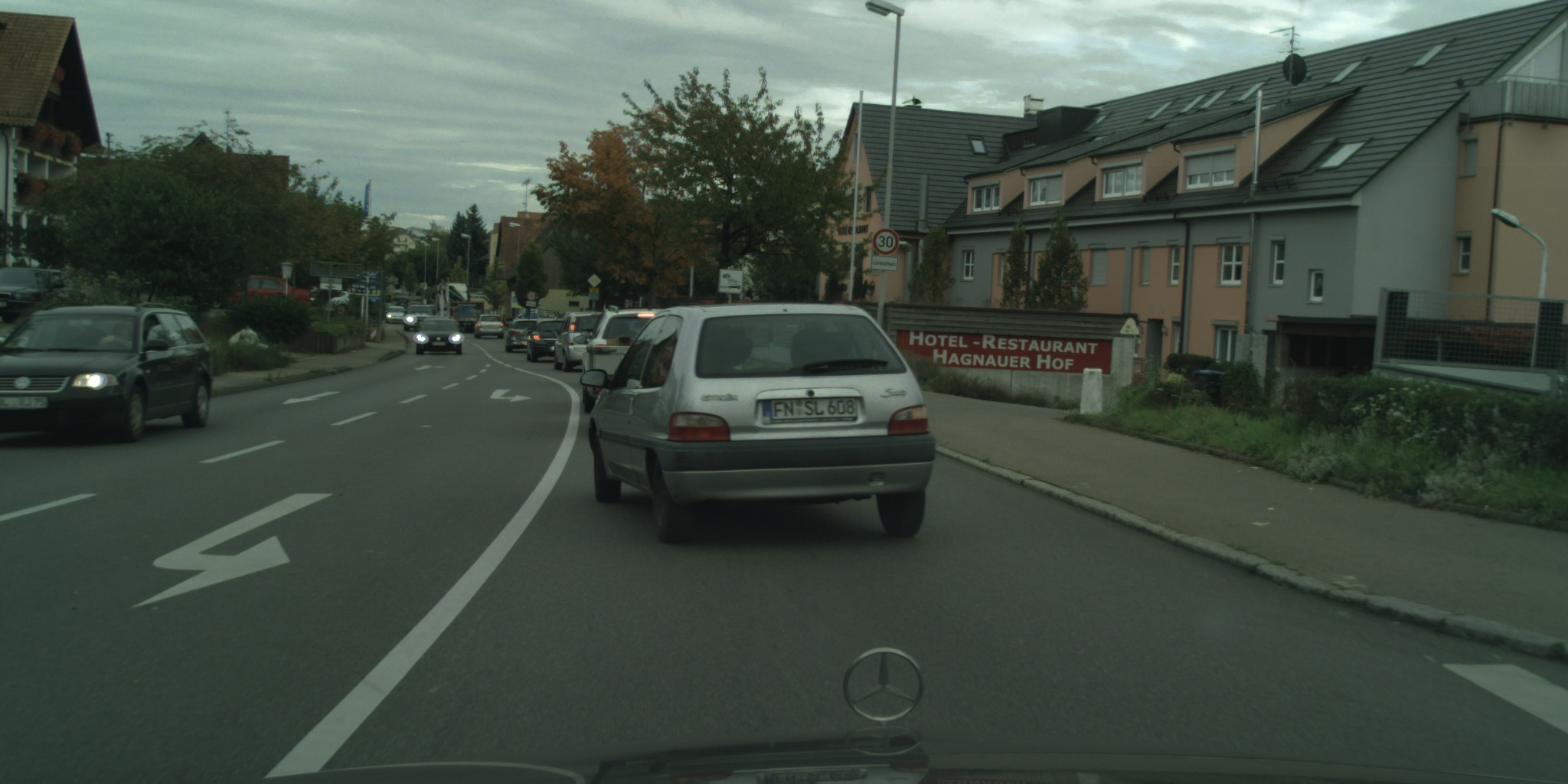} & 
\includegraphics[width=0.2\textwidth, height=1.75cm]{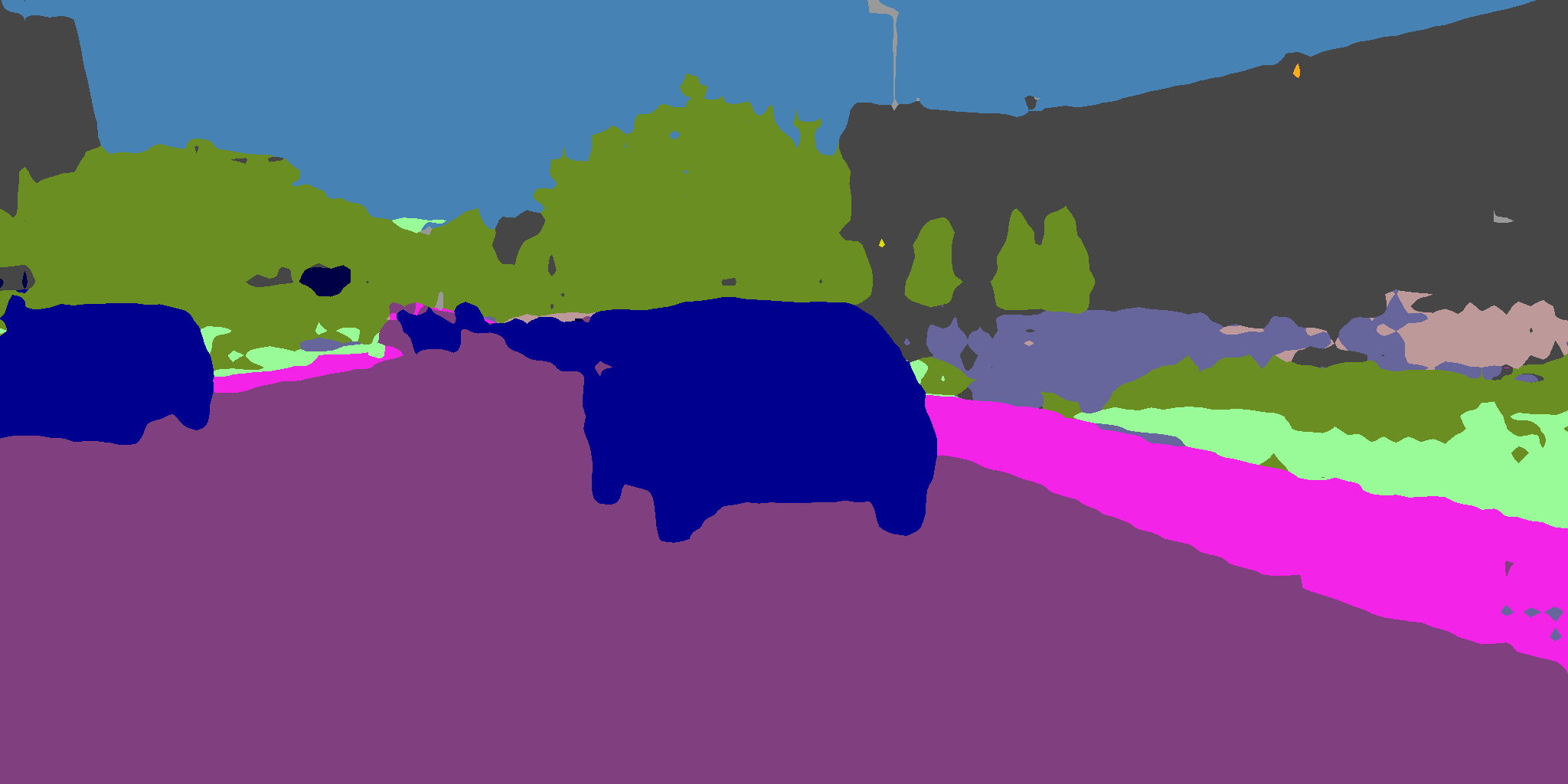} & 
\includegraphics[width=0.2\textwidth, height=1.75cm]{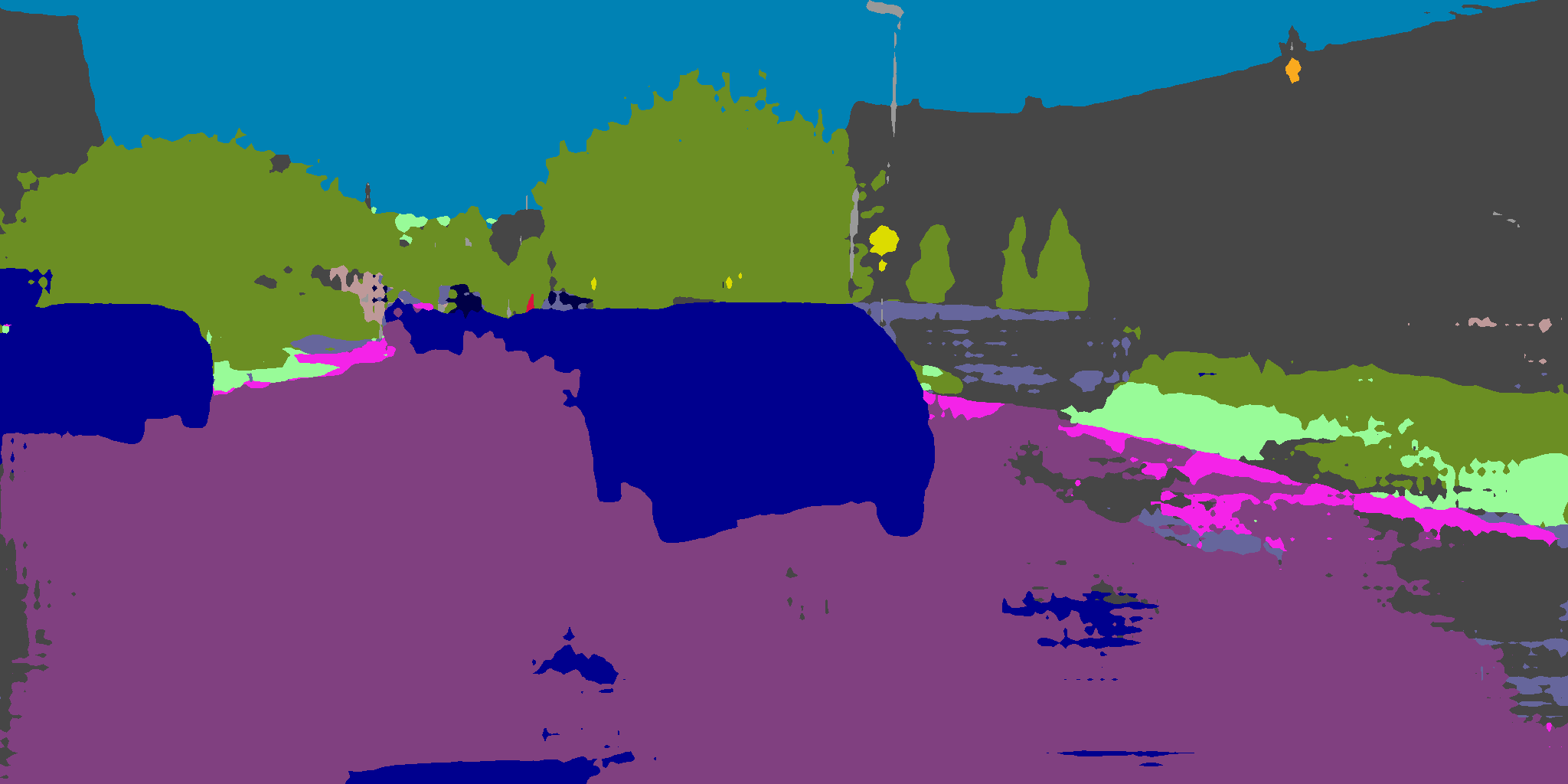} & 
\includegraphics[width=0.2\textwidth, height=1.75cm]{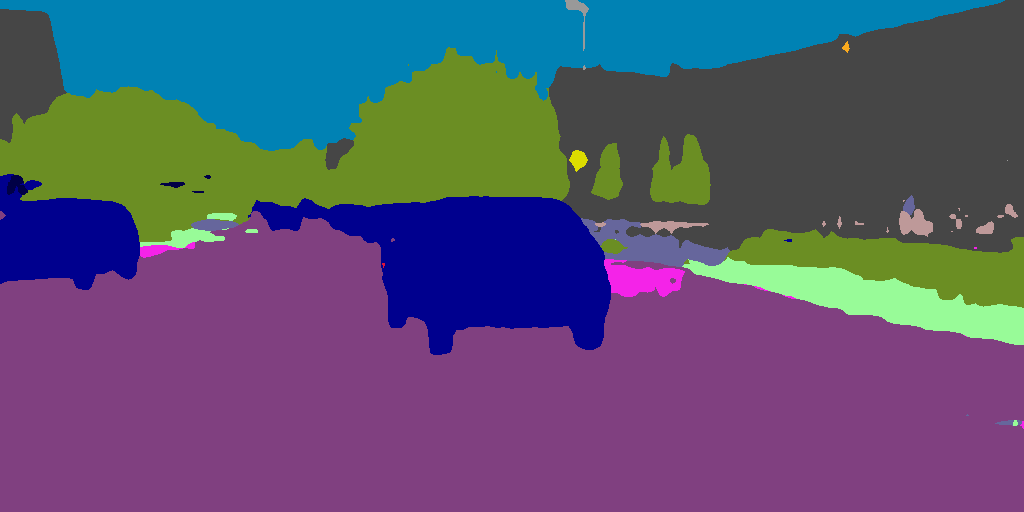} & 
\includegraphics[width=0.2\textwidth, height=1.75cm]{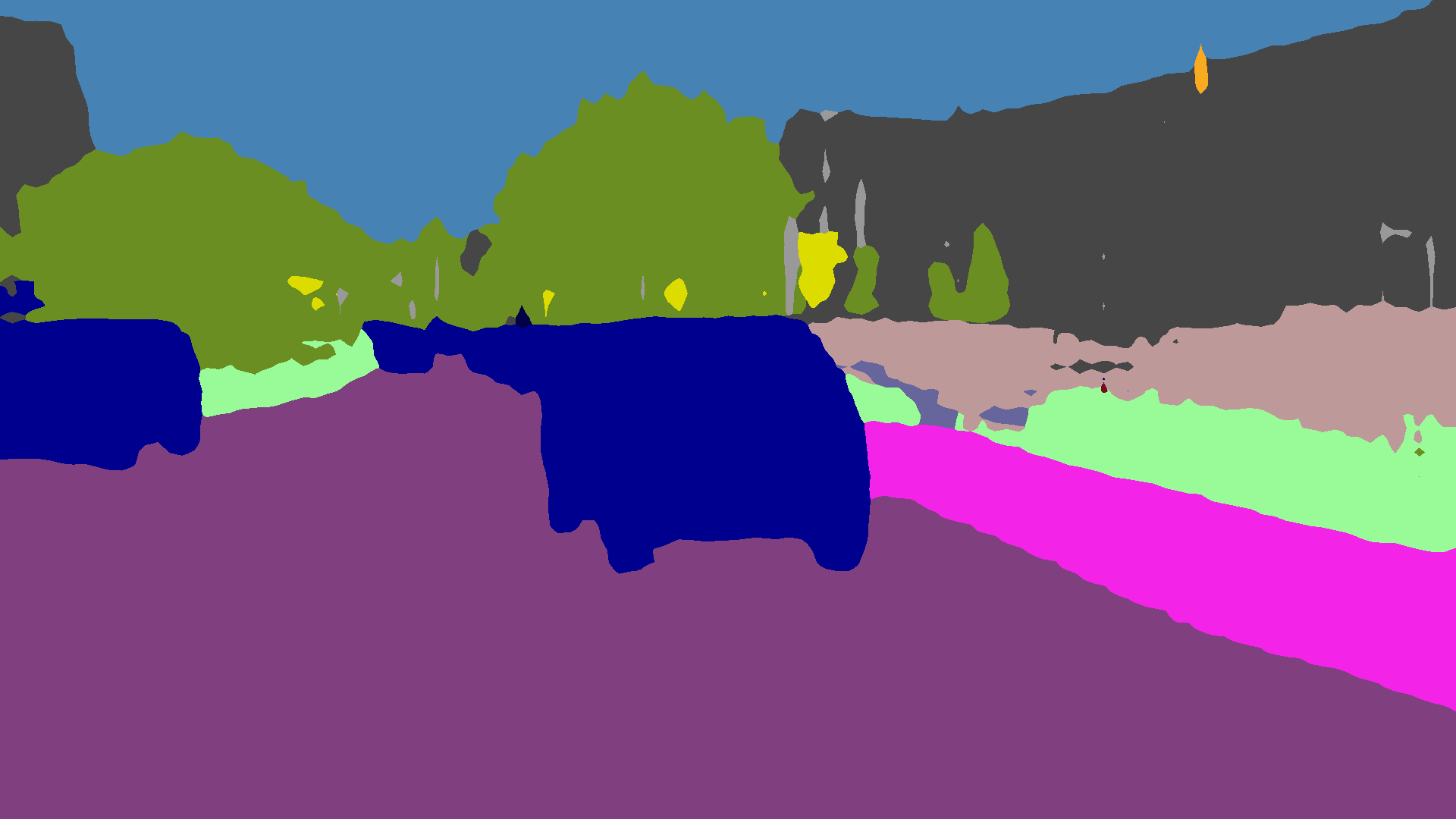} \\
Input & ADVENT & FDA & DACS & DANNet \\
\includegraphics[width=0.2\textwidth, height=1.75cm]{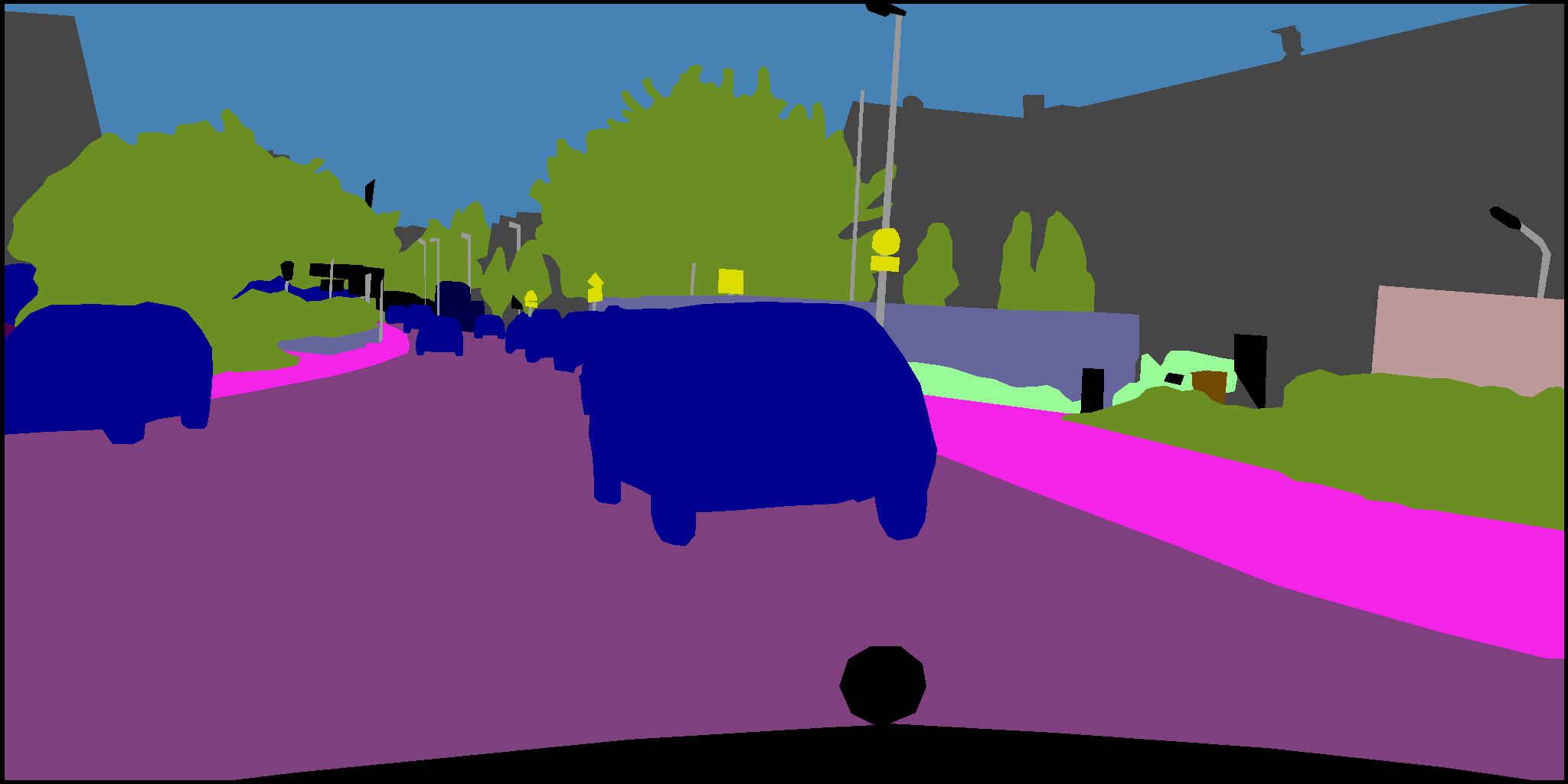} & 
\includegraphics[width=0.2\textwidth, height=1.75cm]{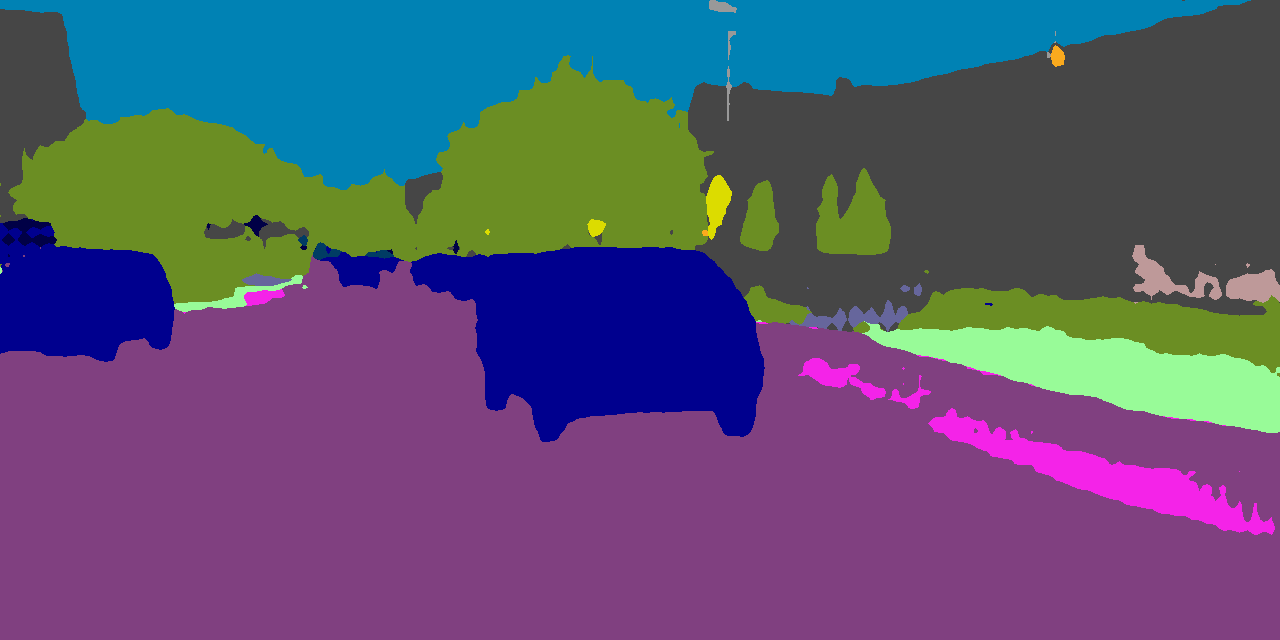} & 
\includegraphics[width=0.2\textwidth, height=1.75cm]{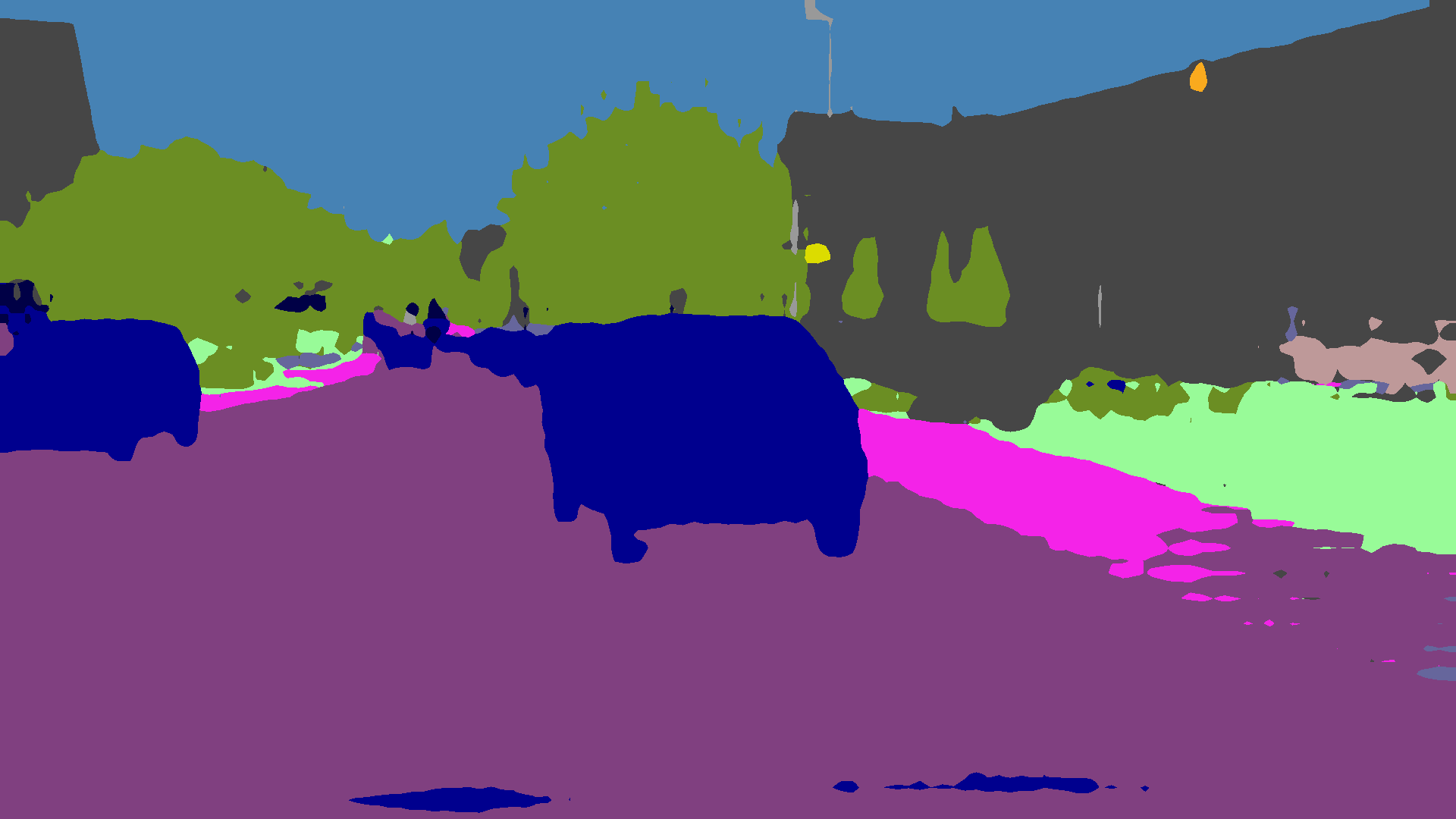} & 
\includegraphics[width=0.2\textwidth, height=1.75cm]{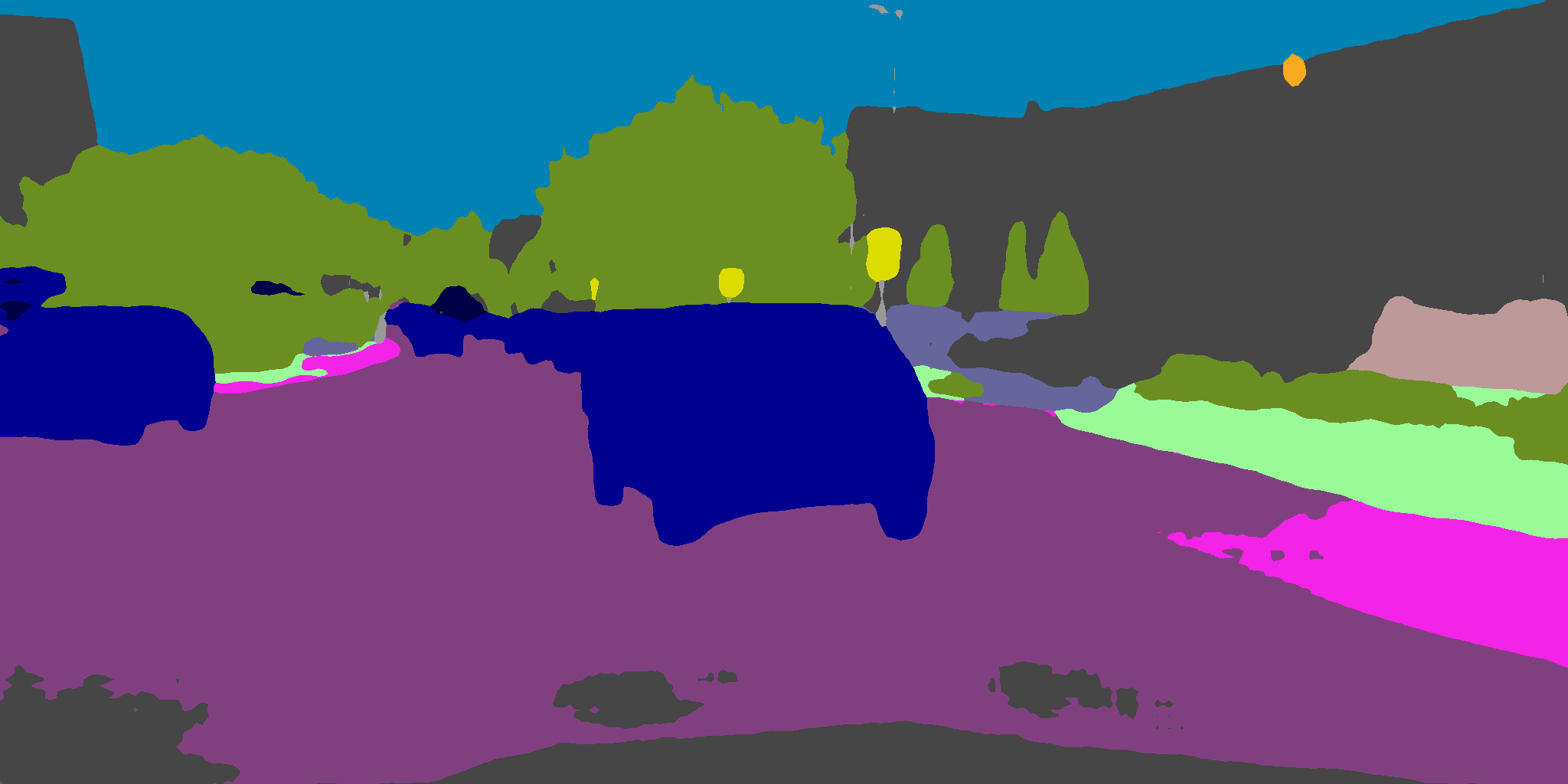} & 
\includegraphics[width=0.2\textwidth, height=1.75cm]{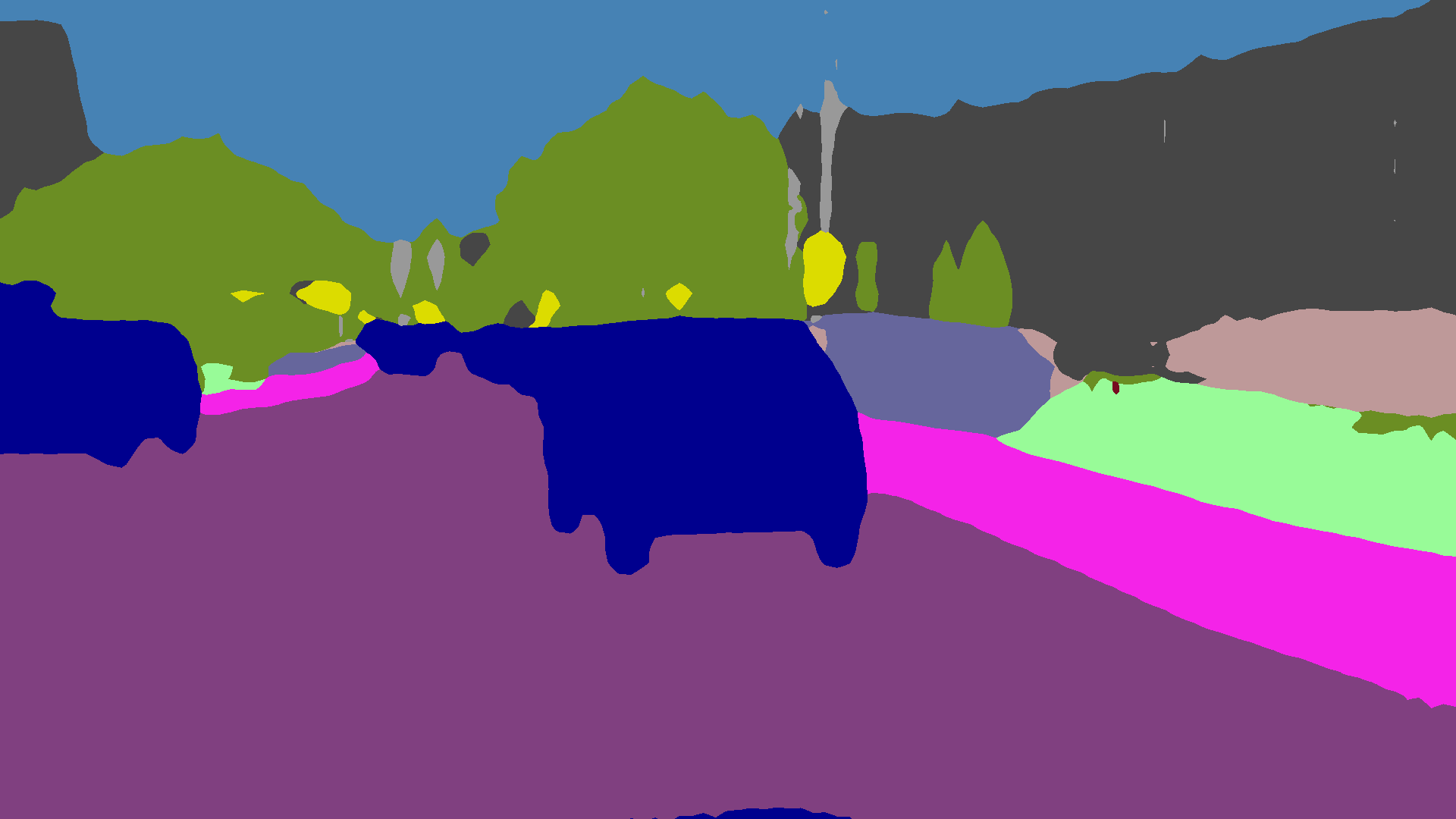} \\
Ground Truth & PixMatch & DISE & ProDA & DGSS \\
\end{tabular}
\end{adjustbox}
\caption{Qualitative Results on Cityscapes Dataset following GTAV $\to$ Cityscapes scenario.}
\label{fig_5}
\vspace{-4mm}
\end{figure*}

\section{Experiments}

\noindent
\textbf{Datasets and Evaluation Metrics :} In order to extensively examine proposed framework, we utilize standard synthetic datasets such as GTA-V \cite{richter2016playing} and SYNTHIA \cite{ros2016SYNTHIA} that act as source domains. Subsequently we use Cityscapes \cite{Cordts2016Cityscapes}, Dark-Zurich \cite{SDV20} and Night-Driving \cite{dai2018dark} as target domains for evaluating the following scenarios SYNTHIA $\to$ Cityscapes, GTAV $\to$ Cityscapes, GTAV $\to$ Dark Zurich and GTAV $\to$ Night Driving. For convenience we summarize the dataset properties in Tab. \ref{tab_1} and elaborate the evaluation scenarios as, 

\begin{table}[thb]
\scriptsize
\centering
\renewcommand{\tabcolsep}{2pt} % adjust horizontal space
\begin{adjustbox}{width=\columnwidth}
\begin{tabular}{lccccc}
\Xhline{3\arrayrulewidth} \noalign{\vskip 1pt}
Dataset & GTAV & SYNTHIA & Cityscapes & Dark-Zurich & Night-Driving \\
\Xhline{2\arrayrulewidth} \noalign{\vskip 1pt}
\# Classes  & 19 & 16 & 19 & 19 & 19 \\
Total Samples  & 24966 & 9400 & 6000 & 8377 & 50 \\
Eval. Samples  & - & - & 500 & 50 & 50 \\
Resolution & 1914 $\times$ 1052 & 1280 $\times$ 760 & 2048 $\times$ 1024 & 1920 $\times$ 1080 & 1920 $\times$ 1080 \\

\Xhline{3\arrayrulewidth} \noalign{\vskip 1pt}
\end{tabular}
\end{adjustbox}
\caption{Summary of different datasets used in this paper.}
\vspace{-4mm}
\label{tab_1}
\end{table}

\begin{itemize}
 \item Consistent with prior works we use the validation subset of the cityscapes dataset for evaluating GTAV $\to$ Cityscapes and SYNTHIA $\to$ Cityscapes scenarios.
 \item To evaluate performance on night conditions, we retrain the SoTA algorithms using night subset of Dark-Zurich dataset and evaluate the performance on night conditions i.e. GTAV $\to$ Dark Zurich.
 \item For examining domain generalization performance of SoTA algorithms we use Night Driving dataset i.e. GTAV $\to$ Night Driving.
\end{itemize}

\noindent 
Since there lacks prior works focusing on achieving domain invariant semantic segmentation performance, we evaluate the performance of proposed framework with SoTA domain adaptation algorithms with ResNet-101 \cite{he2016deep} backbone based DeepLab-v2 \cite{chen2017deeplab} such as ADVENT \cite{vu2018advent}, IAST \cite{mei2020instance}, FDA \cite{yang2020fda}, DA-SAC \cite{Araslanov_DASAC}, DACS \cite{tranheden2021dacs}, DANNet \cite{WU_2021_CVPR},  DISE \cite{chang2019all}, Max. SL \cite{chen2019domain}, MetaCorrection \cite{guo2021metacorrection}, PixMatch \cite{melaskyriazi2021pixmatch}, and ProDA \cite{zhang2021prototypical}. Furthermore for quantitative evaluation we use IoU $(=\frac{TP}{TP + FP +FN})$ and mIOU metrics where TP, FP and FN are abbreviations for true positive, false positive and false negatives respectively and mIOU is calculated by averaging per-class IoU.

\noindent
\textbf{Standard Evaluation with Known Target :} We first compare the performance of SoTA on standard scenarios of GTAV $\to$ Cityscapes and SYNTHIA $\to$ Cityscapes and summarize the performance of Top-5 algorithms in Tab. \ref{tab_2} and Tab. \ref{tab_3} respectively with qualitative results in Fig. \ref{fig_5} for GTAV $\to$ Cityscapes scenario. Owing to space restrictions the complete performance landscape with additional qualitative results and performance of VGG-16 \cite{simonyan2014very} as backbone is included in supplementary. From performance results we can summarize that our proposed algorithm improves performance of baseline by 17.8 mIOU and 23.4 mIoU for GTAV $\to$ Cityscapes and SYNTHIA $\to$ Cityscapes scenarios respectively, consistently improving the classification accuracy for all categories without any prior information to target image or labels. Despite our approach relying solely on source information we achieve competitive performance with respect to SoTA (ProDA) while surpassing it on certain categories, demonstrating that access to target information is not a necessary requirement for performing domain adaptation. Rather we demonstrate that adversarial style mining and subsequent image stylization is sufficient to generate a diverse range of training samples that can be used for achieving domain generalization. 

\begin{table*}[!th]
\scriptsize
\centering
\begin{adjustbox}{width=\textwidth}
\begin{tabular}{l|l|ccccccccccccccccccc|c}
\Xhline{3\arrayrulewidth} \noalign{\vskip 1pt}
\multirow{5}{*}{\rotatebox{90}{Cityscapes $\to$ Dark-Zurich}} & Method  &  
\rotatebox{90}{road} &  
\rotatebox{90}{sidewalk} &  
\rotatebox{90}{building} & 
\rotatebox{90}{wall} & 
\rotatebox{90}{fence} & 
\rotatebox{90}{pole} & 
\rotatebox{90}{light} & 
\rotatebox{90}{sign} & 
\rotatebox{90}{vegetation} & 
\rotatebox{90}{terrain} & 
\rotatebox{90}{sky} & 
\rotatebox{90}{person} & 
\rotatebox{90}{rider} & 
\rotatebox{90}{car} & 
\rotatebox{90}{truck} & 
\rotatebox{90}{bus} & 
\rotatebox{90}{train} & 
\rotatebox{90}{motorbike} & 
\rotatebox{90}{bike} & 
mIoU \\
\Xcline{2-22}{2\arrayrulewidth} \noalign{\vskip 1pt}
& Source Only & 34.1 & 24.9 & 34.8 & 15.1 & 10.7 & 11.9 & 20.1 & 10.2 & 30.7 & 14.3 & 0.0 & 12.8 & 24.6 & 34.9 & 0.0 & 0.0 & 0.0 & 10.1 & 17.2 & 20.4\\
\Xcline{2-22}{2\arrayrulewidth} \noalign{\vskip 1pt}
& DMAda & 65.5 & 29.1 & 48.6 & 21.3 & 14.3 & 10.3 & 26.8 & 19.9 & 39.4 & 13.8 & 0.4 & 43.3 & 50.2 & 69.4 & 0.0 & 0.0 & 0.0 & 22.4 & 10.4 & 25.5 \\
& GCMA  & 69.7 & 46.9 & 58.8 & 22.0 & 20.0 & 12.1 & 20.5 & 21.6 & 54.8 & 31.0 & 32.1 & 23.5 & 47.5 & 55.5 & 0.0 & 0.0 & 0.0 & 29.7 & 21.0 & 29.8 \\
& MGCDA & 69.3 & 49.3 & 66.2 & 7.8 & 11.0 & 14.4 & 28.9 & 23.0 & 44.1 & 18.0 & 27.8 & 22.1 & 53.5 & 54.7 & 0.0 & 0.0 & 0.0 & 29.1 & 22.7 & 28.5 \\
\Xcline{2-22}{2\arrayrulewidth} \noalign{\vskip 1pt}
& DANNet & \underline{88.6 }  & \underline{53.0 }  & \underline{74.3 }  & \underline{29.8 }  & \underline{30.1 }  & \underline{14.5 }  & 21.2& \underline{15.7 }  & \underline{60.9 }  & \underline{21.9 }  & \underline{81.3 }  & 15.9& 12.1& \underline{36.0 }  & 0.0 & 0.0 & 0.0 & \underline{26.7 }  & \underline{27.0 }  & \underline{32.0 }\\
& MetaCorrection & 14.9& 1.8 & 34.5& 2.6 & 5.6 & 14.4& 8.5 & 3.4 & 31.6& 2.5 & 0.0 & 2.5 & 0.0 & 1.5 & 0.0 & 0.0 & 0.0 & 0.0 & 0.0 & 6.5  \\
& PixMatch & 43.4& 9.5 & 37.9& 4.5 & 7.4 & 3.3 & 14.1& 5.6 & 22.2& 3.1 & 0.1 & 15.8& \underline{21.5 }  & 2.6 & 0.0 & 0.0 & 0.0 & 11.2& 0.0 & 10.6 \\
& ProDA  & 48.7& 11.7& 45.1& 7.0 & 18.9& 8.7 & 20.3& 5.1 & 31.3& 18.6& 2.4 & \underline{18.7 }  & 0.0 & 7.4 & 0.0 & 0.0 & 0.0 & 0.0 & 4.9 & 13.1 \\
& DISE& 52.0& 0.7 & 36.1& 4.4 & 2.2 & 4.3 & \underline{24.6 }  & 3.7 & 28.0& 4.7 & 2.9 & 16.7& 1.6 & 4.5 & 0.0 & 0.0 & 0.0 & 8.2 & 0.0 & 10.2 \\
& DGSS& \textbf{92.4 } & \textbf{64.8 } & \textbf{46.9 } & \textbf{31.8 } & \textbf{25.9 } & \textbf{18.0 } & \textbf{38.2 } & \textbf{15.2 } & \textbf{61.8 } & \textbf{26.8 } & \textbf{80.4 } & \textbf{30.0 } & \textbf{33.8 } & \textbf{54.0 } & 0.0 & 0.0 & 0.0 & \textbf{32.1 } & \textbf{42.2 } & \textbf{36.5 }  \\

\Xhline{3\arrayrulewidth} \noalign{\vskip 1pt}
\end{tabular}
\end{adjustbox}
\caption{Performance Comparison with different SoTA domain adaptation algorithms for Cityscapes $\to$ Dark-Zurich scenario.}
\label{tab_4}
\end{table*}

\begin{figure*}[!th]
\renewcommand{\tabcolsep}{1pt} % adjust horizontal space
\renewcommand{\arraystretch}{1} % adjust vertical space
\scriptsize
\centering
\begin{adjustbox}{width=\textwidth}
\begin{tabular}{ccccc}

% GOPR0356_frame_000360_rgb_anon.png
\includegraphics[width=0.2\textwidth, height=1.75cm]{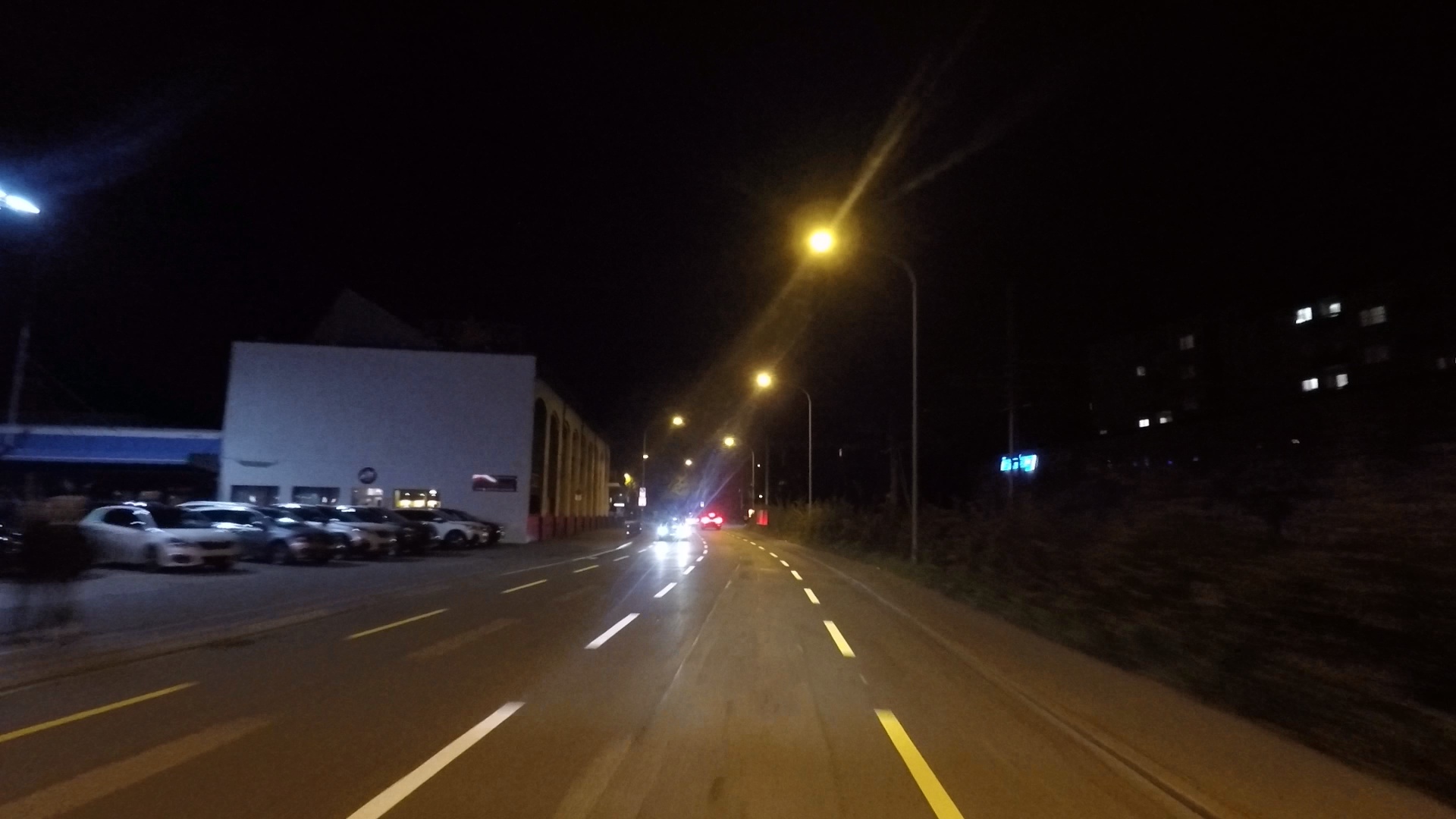} & 
\includegraphics[width=0.2\textwidth, height=1.75cm]{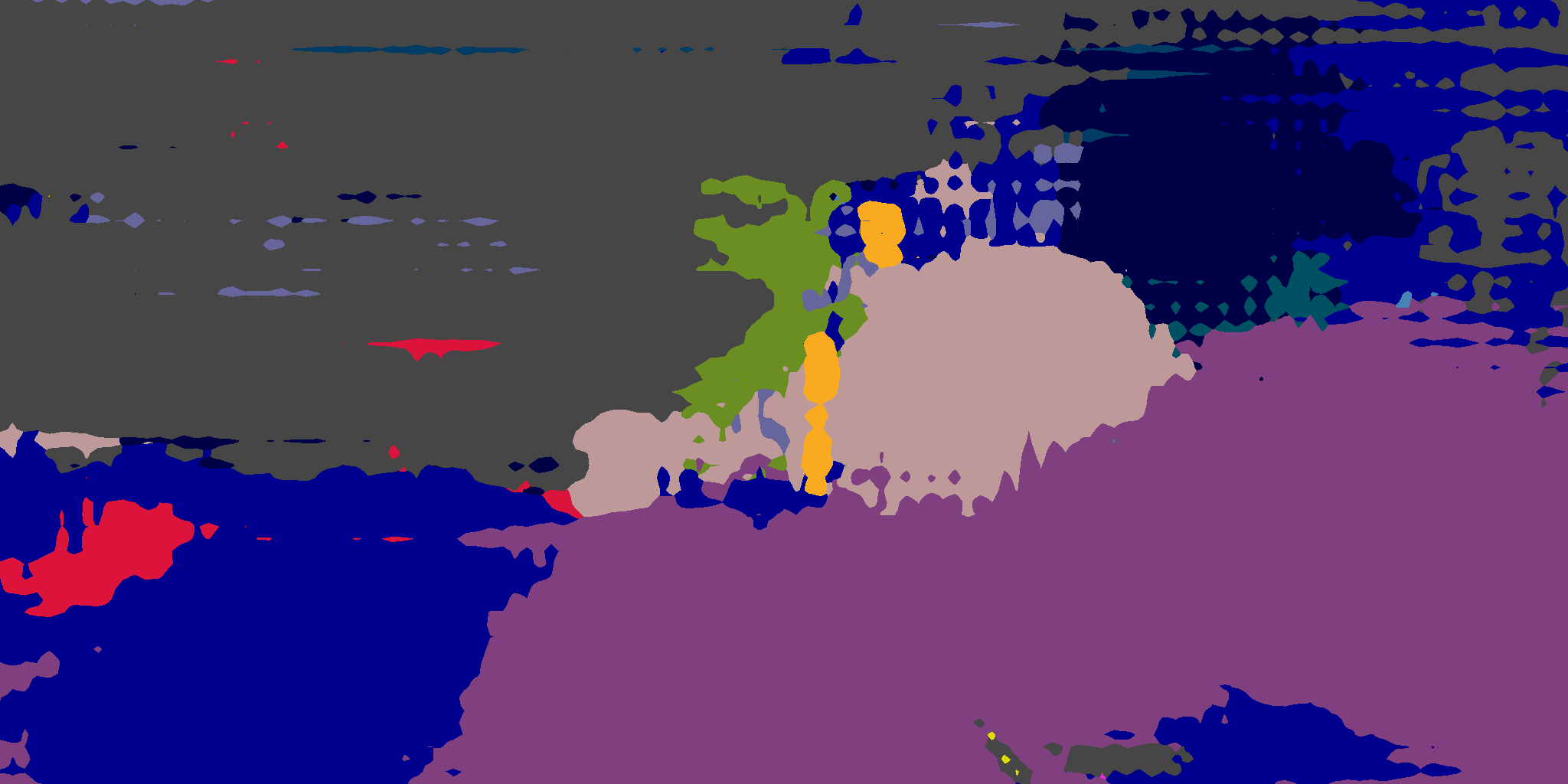} & 
\includegraphics[width=0.2\textwidth, height=1.75cm]{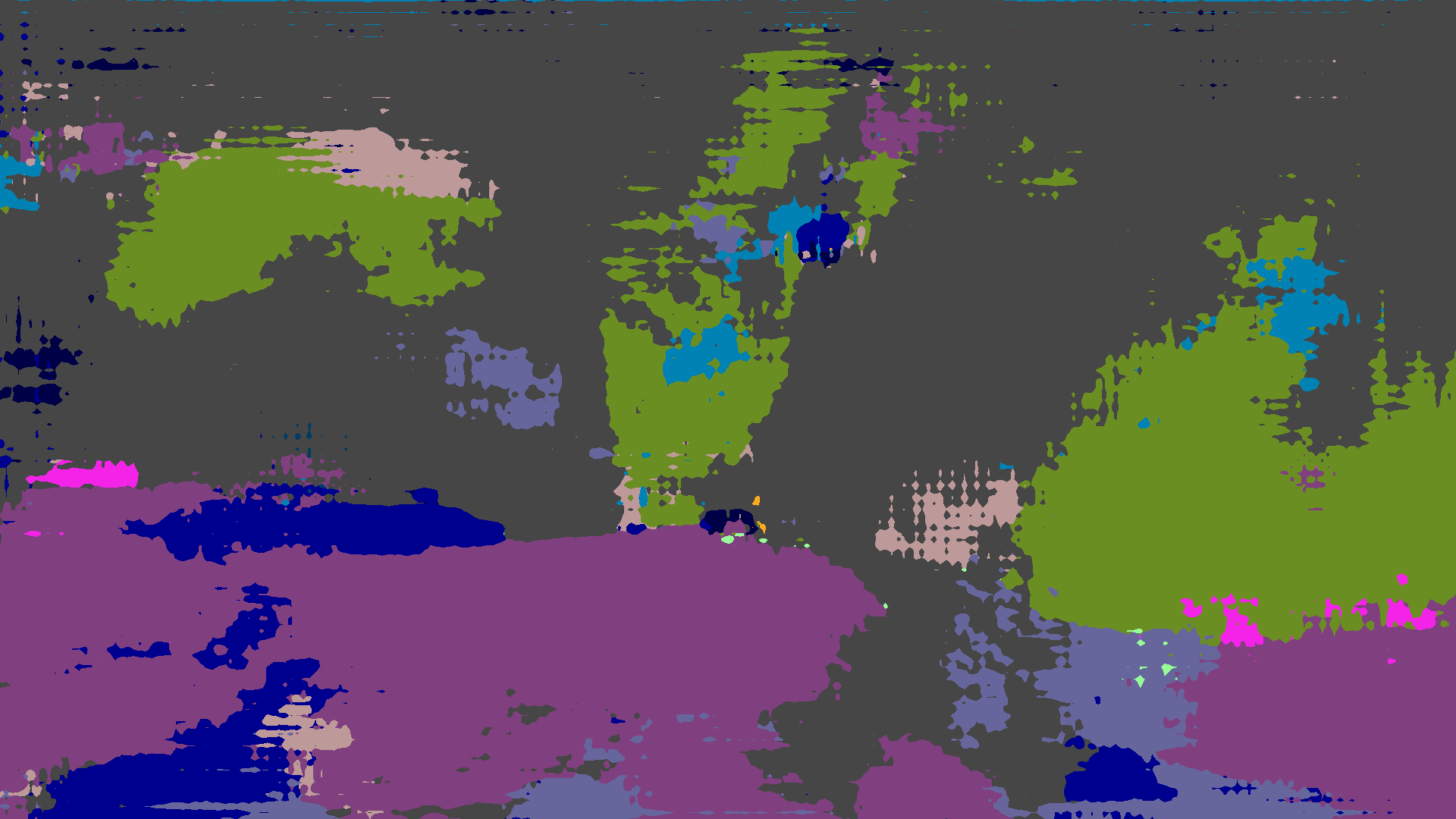} & 
\includegraphics[width=0.2\textwidth, height=1.75cm]{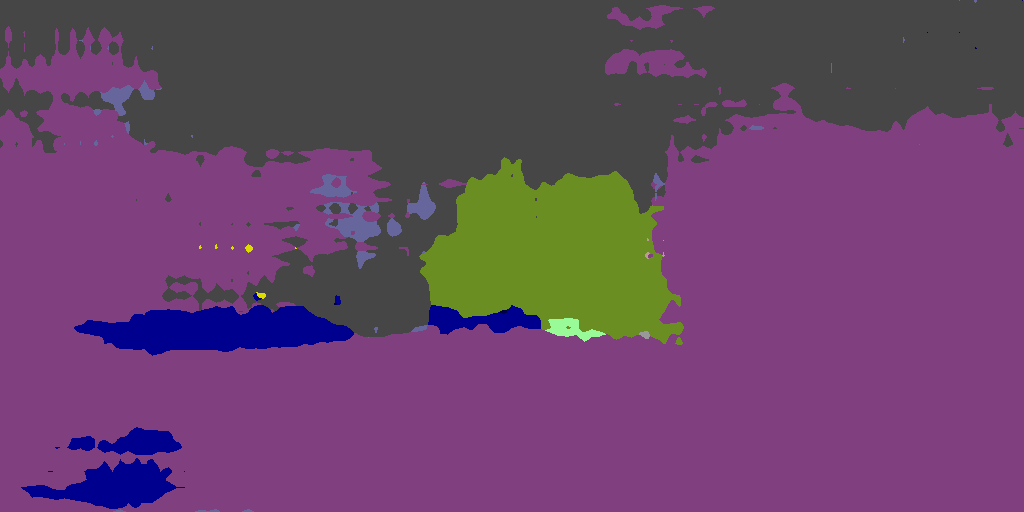} & 
\includegraphics[width=0.2\textwidth, height=1.75cm]{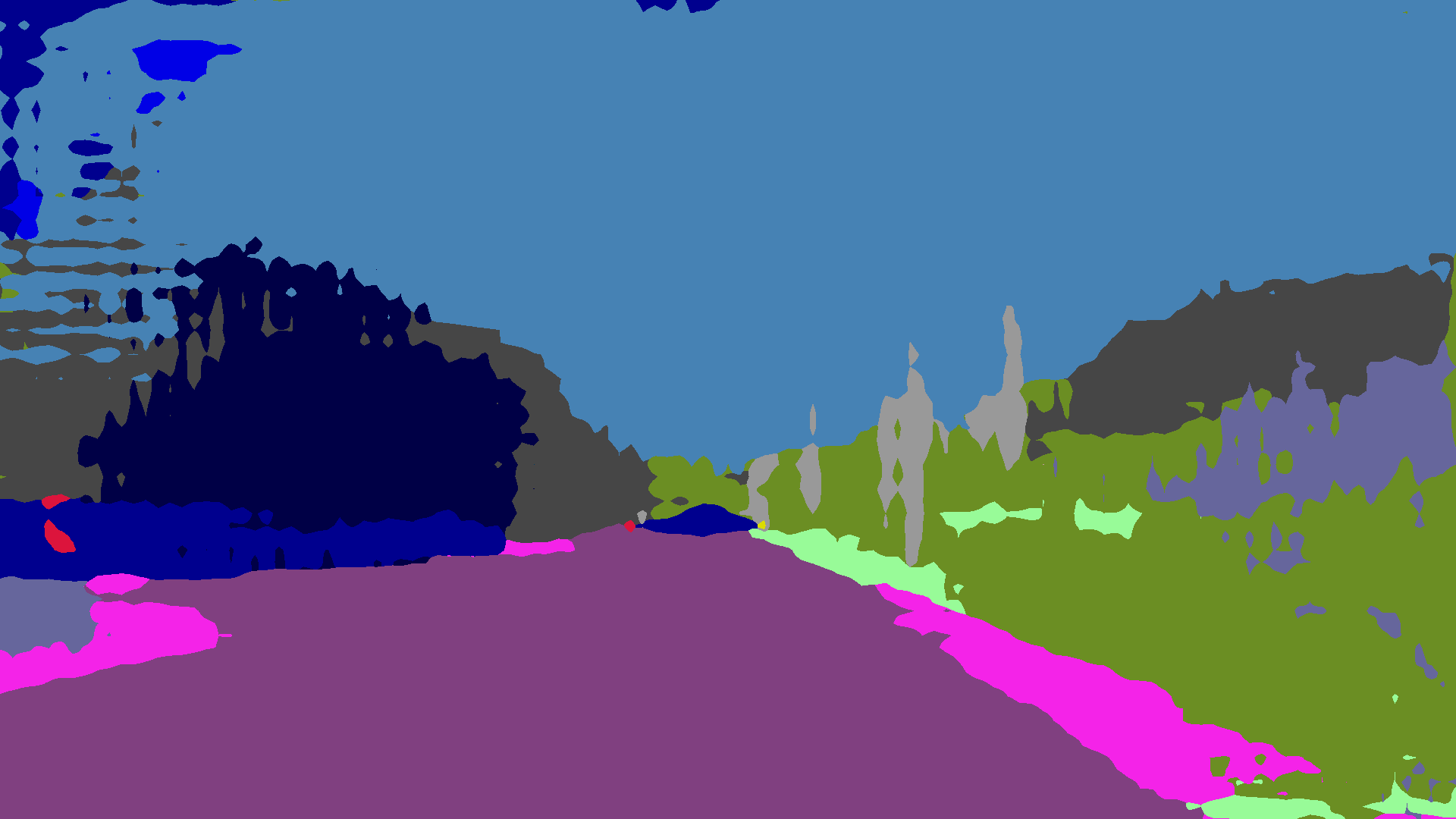} \\
Input & ADVENT & FDA & DACS & DANNet \\
\includegraphics[width=0.2\textwidth, height=1.75cm]{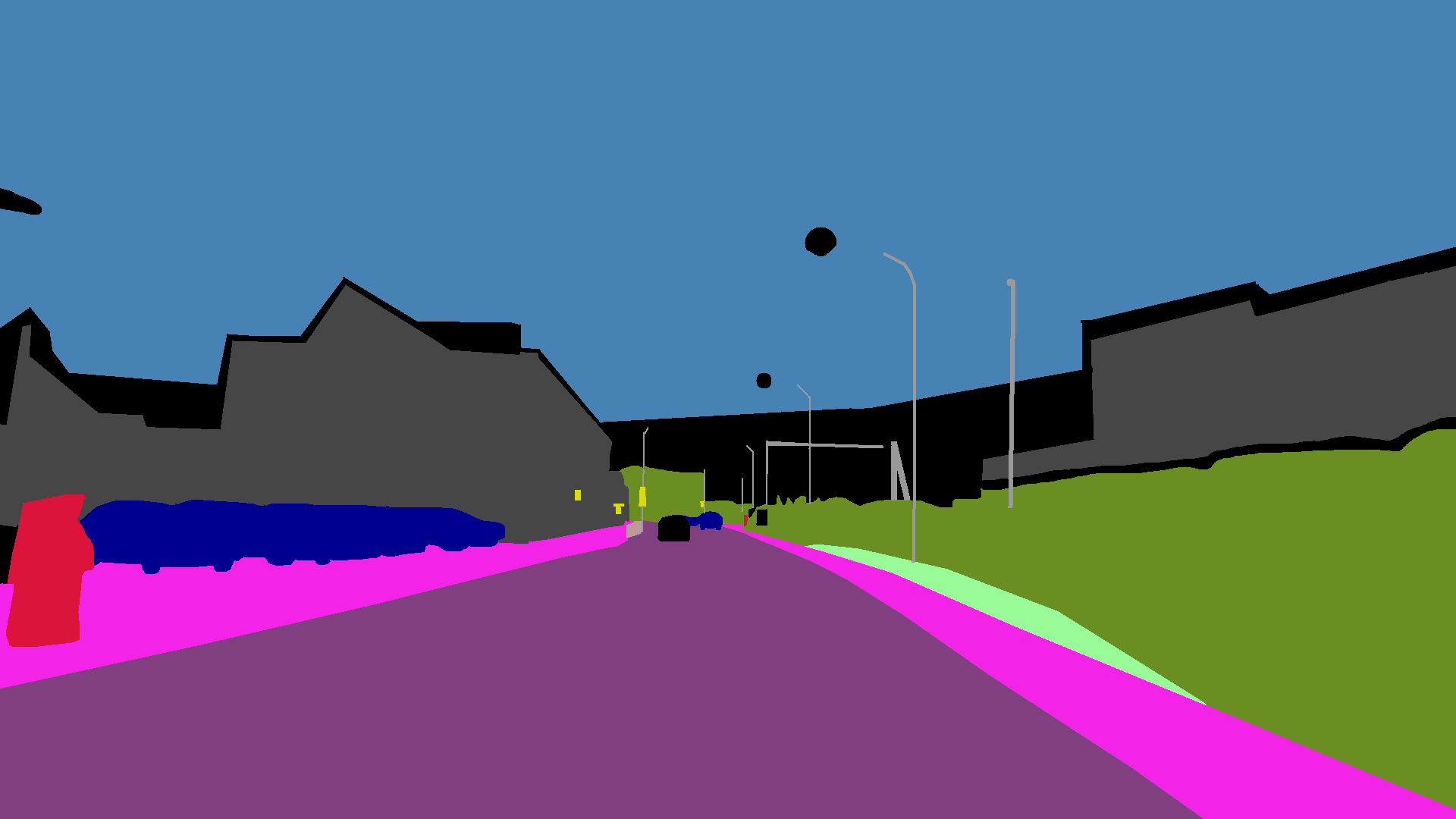} & 
\includegraphics[width=0.2\textwidth, height=1.75cm]{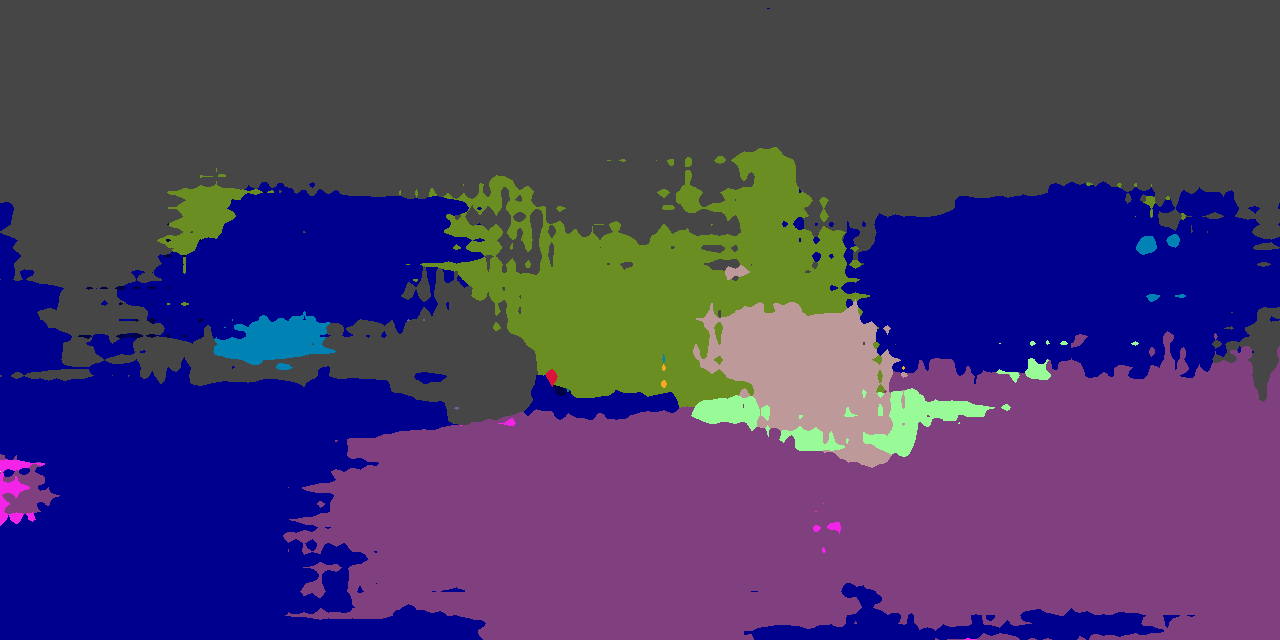} & 
\includegraphics[width=0.2\textwidth, height=1.75cm]{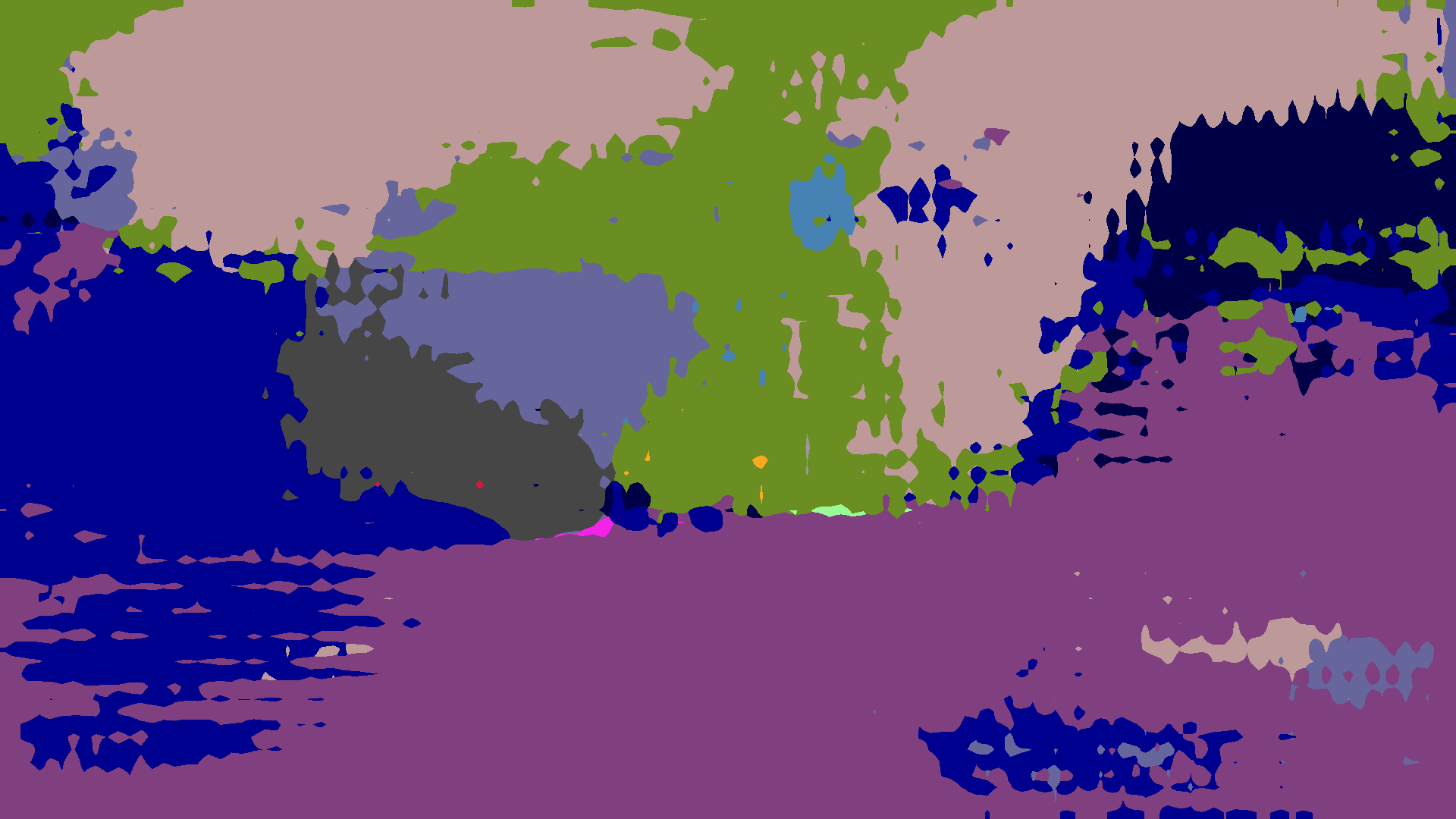} & 
\includegraphics[width=0.2\textwidth, height=1.75cm]{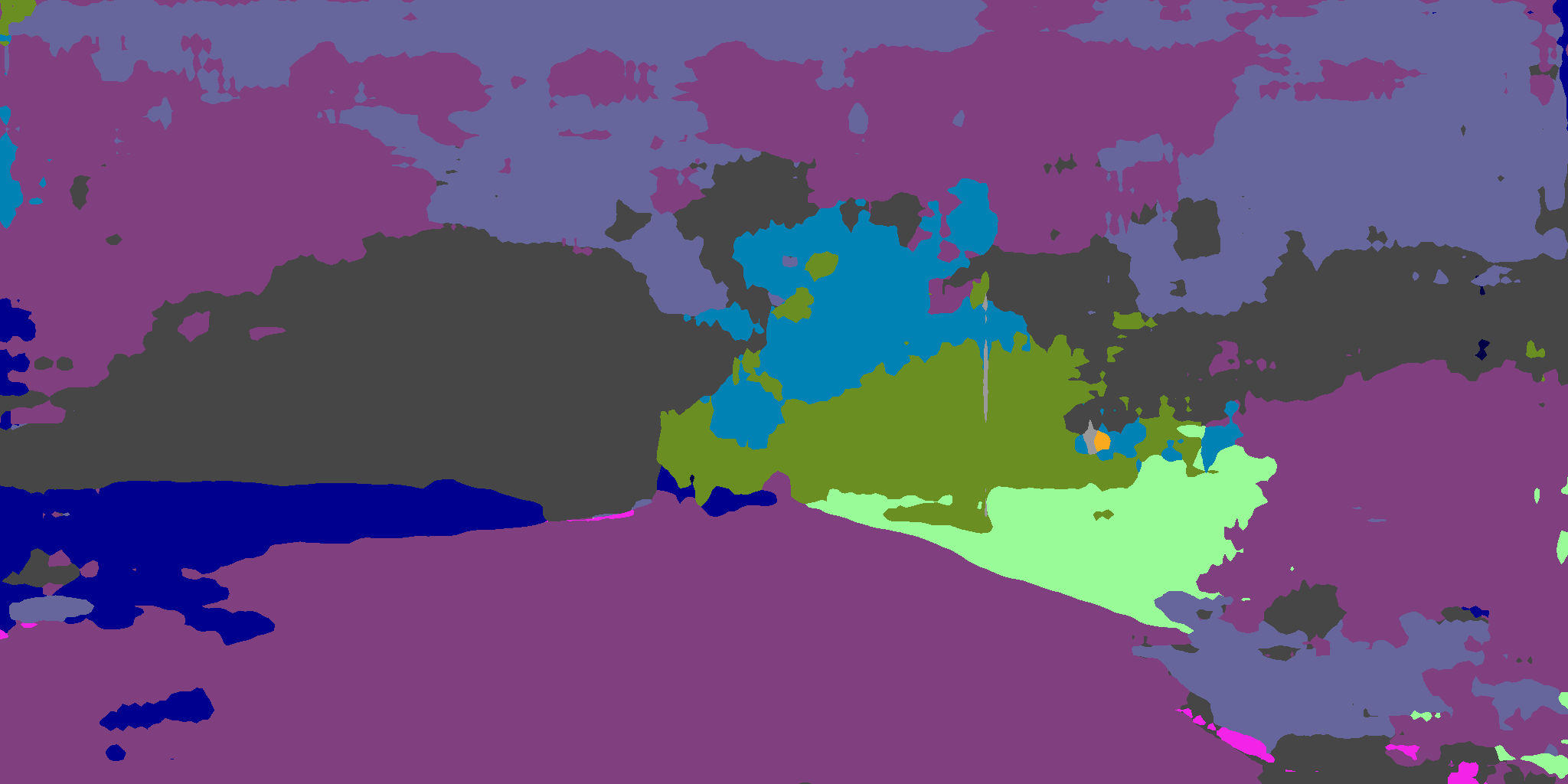} & 
\includegraphics[width=0.2\textwidth, height=1.75cm]{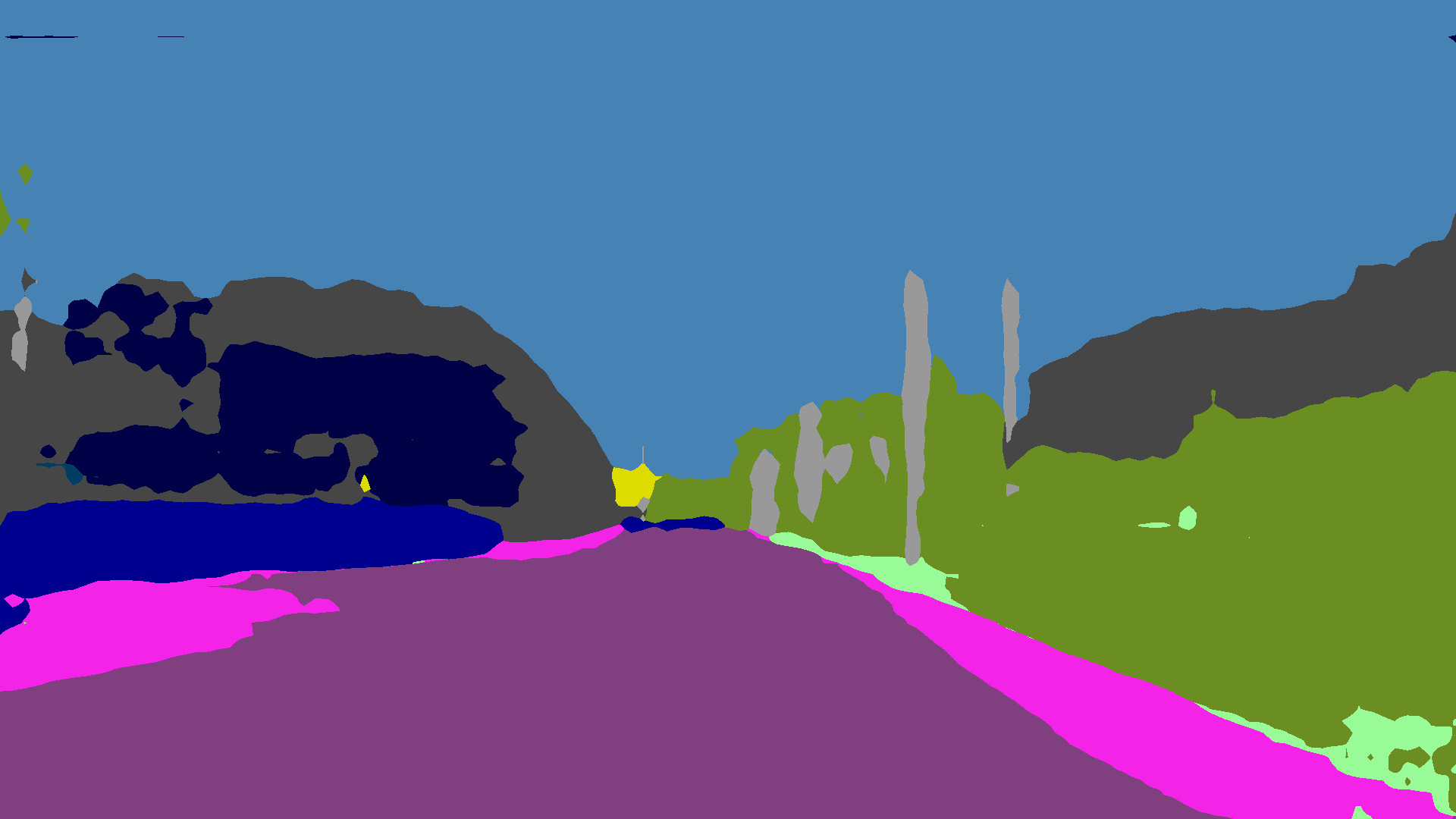} \\
Ground Truth & PixMatch & DISE & ProDA & DGSS \\
\end{tabular}
\end{adjustbox}
\caption{Qualitative Results on Dark-Zurich Dataset following Cityscapes $\to$ Dark-Zurich scenario.}
\label{fig_6}
\vspace{-4mm}
\end{figure*}

\noindent
\textbf{Known Target but different Illumination :} Majority of prior works while evaluating performance on known target domain overlook the effect of illumination on segmentation performance. Hence we additionally evaluate performance under varying illumination conditions and thus utilize Dark-Zurich dataset that captures images of a scene at different illumination conditions i.e. day, twilight and night using same camera with day images acting as guidance for annotating night images. Hence, we retrain the top performing SoTA from above using night images, following prior works \cite{sakaridis2020map, sakaridis2019guided, dai2018dark} to ensure SoTA is able to minimize the domain gap, and subsequently evaluate the trained models on the validation set of Dark-Zurich dataset comprising of 50 images. While retraining we follow the individual training process of each algorithm without any modifications. Furthermore, we donot make any modifications to our SS algorithm and summarize the performance in Tab. \ref{tab_4} with qualitative results in Fig. \ref{fig_6}. In addition, we also summarize performance of SoTA on Dark-Zurich dataset i.e. MGCDA \cite{sakaridis2020map}, GCMA \cite{sakaridis2019guided}, DMAda \cite{dai2018dark} that uses RefineNet \cite{lin2017refinenet} as baseline and perform adaption using cityscapes as source domain with night images within Dark-Zurich training set as target domain. However due to different backbone a direct comparison of these algorithms would not be fair, hence we simply include performance of these algorithms to provide quantitative representation of SoTA on Dark-Zurich dataset. From Tab. \ref{tab_4} we can conclude that our proposed framework surpasses previous SoTA (DANNet) by 4.5 mIOU, however it should be noted that while DANNet achieves 32.0 mIoU, it relies on additional relight module integrated within the CNN to ensures illumination consistency irrespective of source image, resulting in higher accuracy. Despite this our proposed framework overcoming the SoTA with re-lightening module demonstrates the robustness ensured by proposed training mechanism. Furthermore from the visual results we concur DGGS to provide a high quality segmentation whereas prior SoTA such as ProDA or DANNet suffer from noisy prediction. We conjecture that utilizing different stylized images enables the underlying network to extract illumination invariant features i.e. the texture of an image and subsequently generate pixel-wise classification with reasonable accuracy, whereas all top performing SoTA are sensitive towards these varying illumination conditions.

\begin{table*}[!th]
\scriptsize
\centering
\begin{adjustbox}{width=\textwidth}
\begin{tabular}{l|l|ccccccccccccccccccc|c}
\Xhline{3\arrayrulewidth} \noalign{\vskip 1pt}
\multirow{7}{*}{\rotatebox{90}{Cityscapes $\to$ Dark Zurich}} & Method  &  
\rotatebox{90}{road} &  
\rotatebox{90}{sidewalk} &  
\rotatebox{90}{building} & 
\rotatebox{90}{wall} & 
\rotatebox{90}{fence} & 
\rotatebox{90}{pole} & 
\rotatebox{90}{light} & 
\rotatebox{90}{sign} & 
\rotatebox{90}{vegetation} & 
\rotatebox{90}{terrain} & 
\rotatebox{90}{sky} & 
\rotatebox{90}{person} & 
\rotatebox{90}{rider} & 
\rotatebox{90}{car} & 
\rotatebox{90}{truck} & 
\rotatebox{90}{bus} & 
\rotatebox{90}{train} & 
\rotatebox{90}{motorbike} & 
\rotatebox{90}{bike} & 
mIoU \\
\Xcline{2-22}{2\arrayrulewidth} \noalign{\vskip 1pt}
& Source Only & 59.0 & 21.8 & 53.0 & 13.3 & 0.0 & 22.5 & 20.2 & 22.1 & 43.5 & 10.4 & 18.0 & 27.4 & 23.8 & 24.1 & 6.4 & 0.0 & 32.3 & 3.4 & 7.4 & 21.5 \\
\Xcline{2-22}{2\arrayrulewidth} \noalign{\vskip 1pt}
& DISE& 64.1 & 3.8& 51.2 & 2.4& 0.1 & 11.2 & 50.5 & 32.1 & 40.1 & 0.0  & 9.0& 39.4 & 0.0& 12.5 & 0.0 & 11.2 & 0.0& 0.0 & 0.0& 17.2 \\
& PixMatch  & 52.0 & 12.6 & 76.6 & 3.3& \textbf{0.5 } & 15.3 & 60.8 & 33.5 & 48.2 & 0.0  & 0.0& 47.3 & 0.0& 9.9& 0.0 & 15.1 & 0.4& 0.0 & 3.5& 19.9 \\
& DA-SAC& 73.4 & 23.0 & 61.8 & 10.6 & 0.0 & 23.5 & 56.7 & 42.7 & 40.5 & 0.0  & 0.1& 49.3 & 0.0& 51.8 & 0.1 & 25.9 & 9.3& 0.0 & \underline{44.7 }  & 27.0 \\
& DANNet & \underline{\textbf{89.7 }} & \underline{57.5 }  & \underline{84.5 }  & \underline{21.0 }  & 0.0 & \underline{35.8 }  & 61.9 & \underline{68.4 }  & \underline{59.0 }  & 0.0  & \underline{50.5 }  & 52.4 & \underline{19.6 }  & \underline{67.6 }  & 0.0 & \underline{70.8 }  & \underline{68.9 }  & 0.0 & 30.5 & \underline{44.1 }  \\
& ProDA & 67.8 & 13.6 & 60.7 & 6.9& \underline{0.2 }  & 17.6 & \underline{63.1 }  & 53.4 & 54.7 & 0.0  & 13.1 & \underline{58.2 }  & 9.8& 22.0 & \textbf{5.6 } & 26.6 & 0.0& \textbf{0.3 } & 36.4 & 26.8 \\
& DGSS& \underline{86.5 }& \textbf{60.2 } & \textbf{86.4 } & \textbf{25.0 } & 0.0 & \textbf{47.7 } & \textbf{72.0 } & \textbf{74.6 } & \textbf{64.8 } & 0.0  & \textbf{56.7 } & \textbf{66.2 } & \textbf{53.6 } & \textbf{75.5 } & \underline{0.4 }  & \textbf{74.2 } & \textbf{74.5 } & 0.0 & \textbf{52.9 } & \textbf{51.1}  \\

\Xhline{3\arrayrulewidth} \noalign{\vskip 1pt}
\end{tabular}
\end{adjustbox}
\caption{Performance Comparison with different SoTA domain adaptation algorithms for Cityscapes $\to$ Dark Zurich evaluated on Night Driving.}
\label{tab_5}
\end{table*}

\begin{figure*}[!th]
\renewcommand{\tabcolsep}{1pt} % adjust horizontal space
\renewcommand{\arraystretch}{1} % adjust vertical space
\scriptsize
\centering
\begin{adjustbox}{width=\textwidth}
\begin{tabular}{cccccc}
% 0_frame_0205_leftImg8bit.png
\includegraphics[width=0.2\textwidth, height=1.75cm]{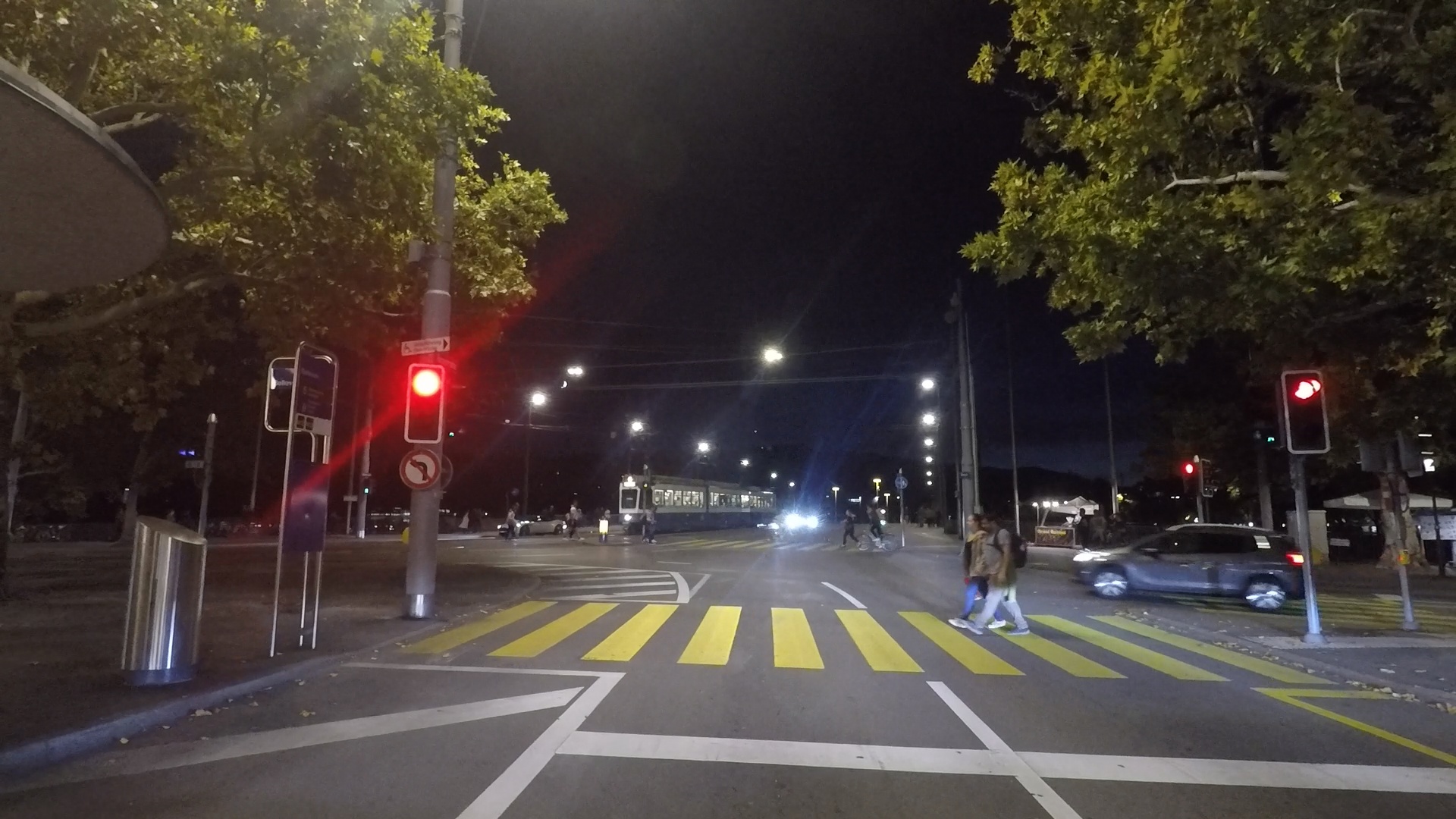} & 
\includegraphics[width=0.2\textwidth, height=1.75cm]{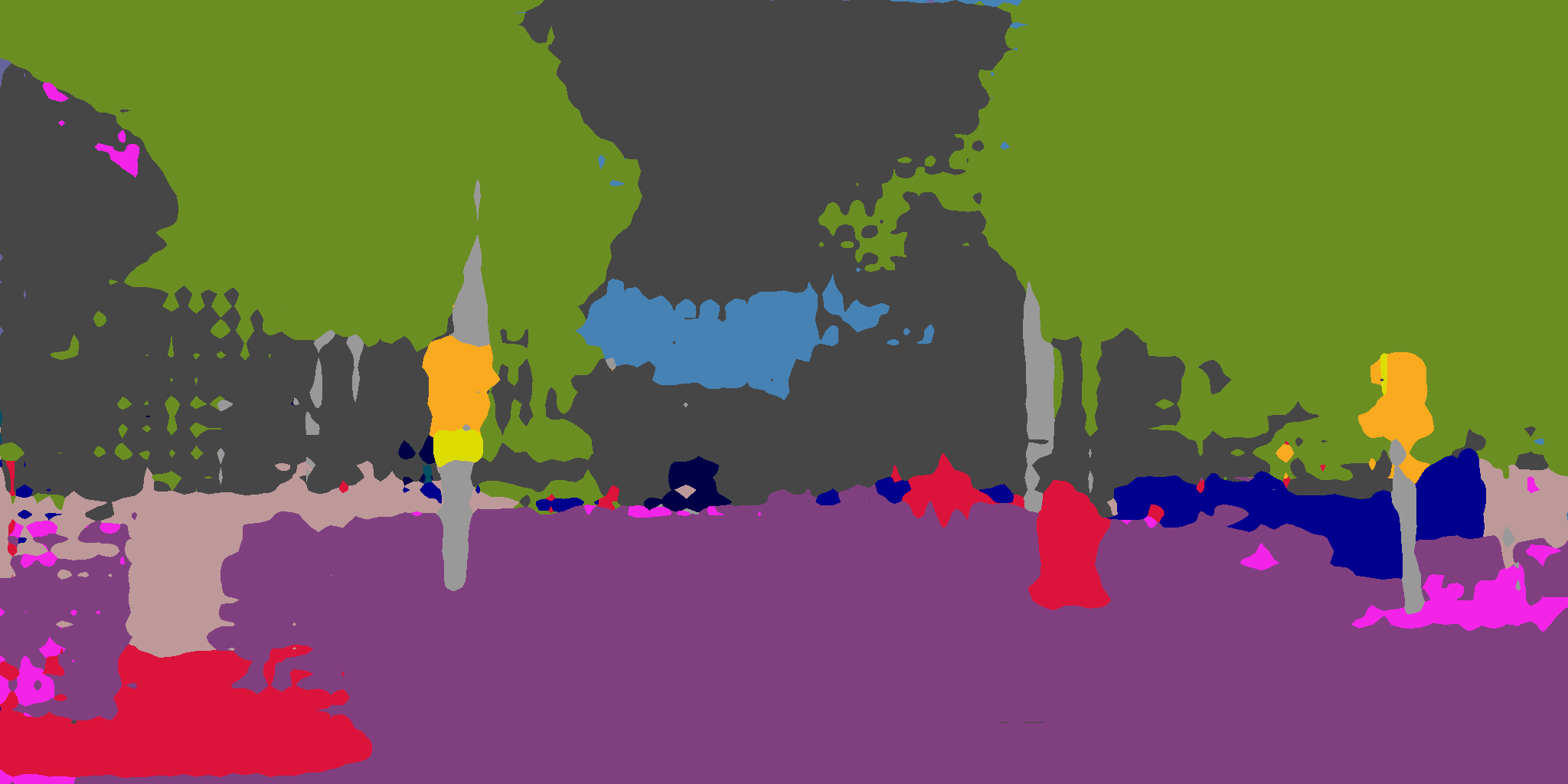} & 
\includegraphics[width=0.2\textwidth, height=1.75cm]{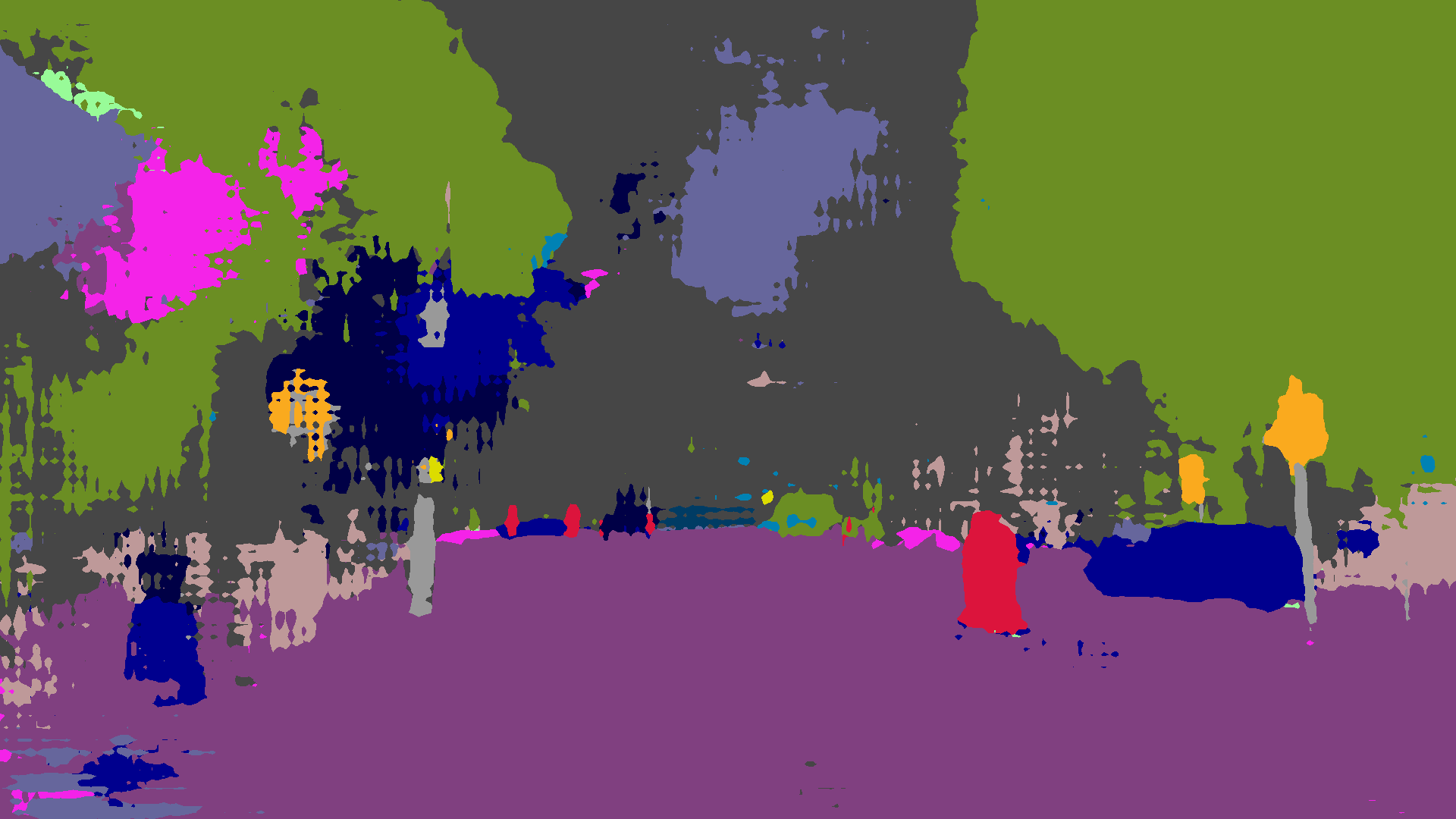} & 
\includegraphics[width=0.2\textwidth, height=1.75cm]{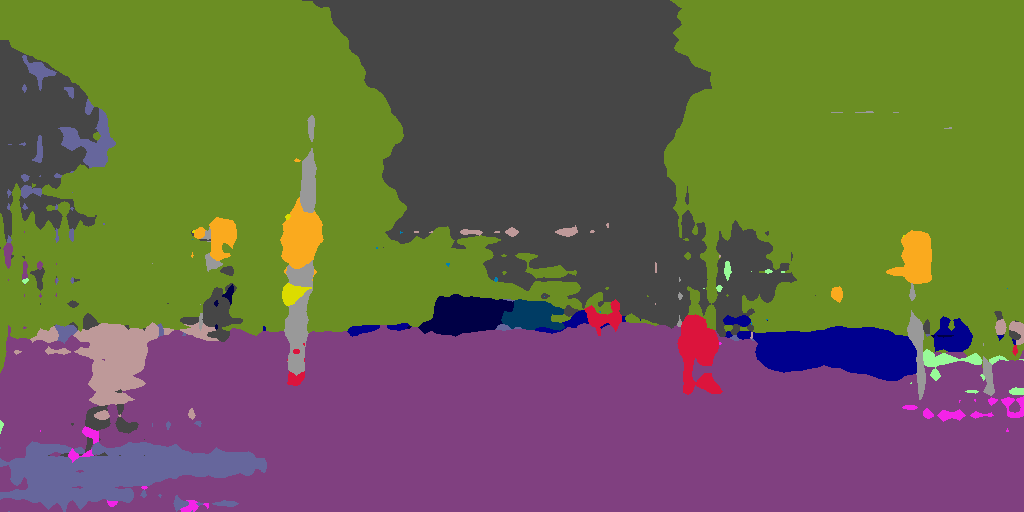} & 
\includegraphics[width=0.2\textwidth, height=1.75cm]{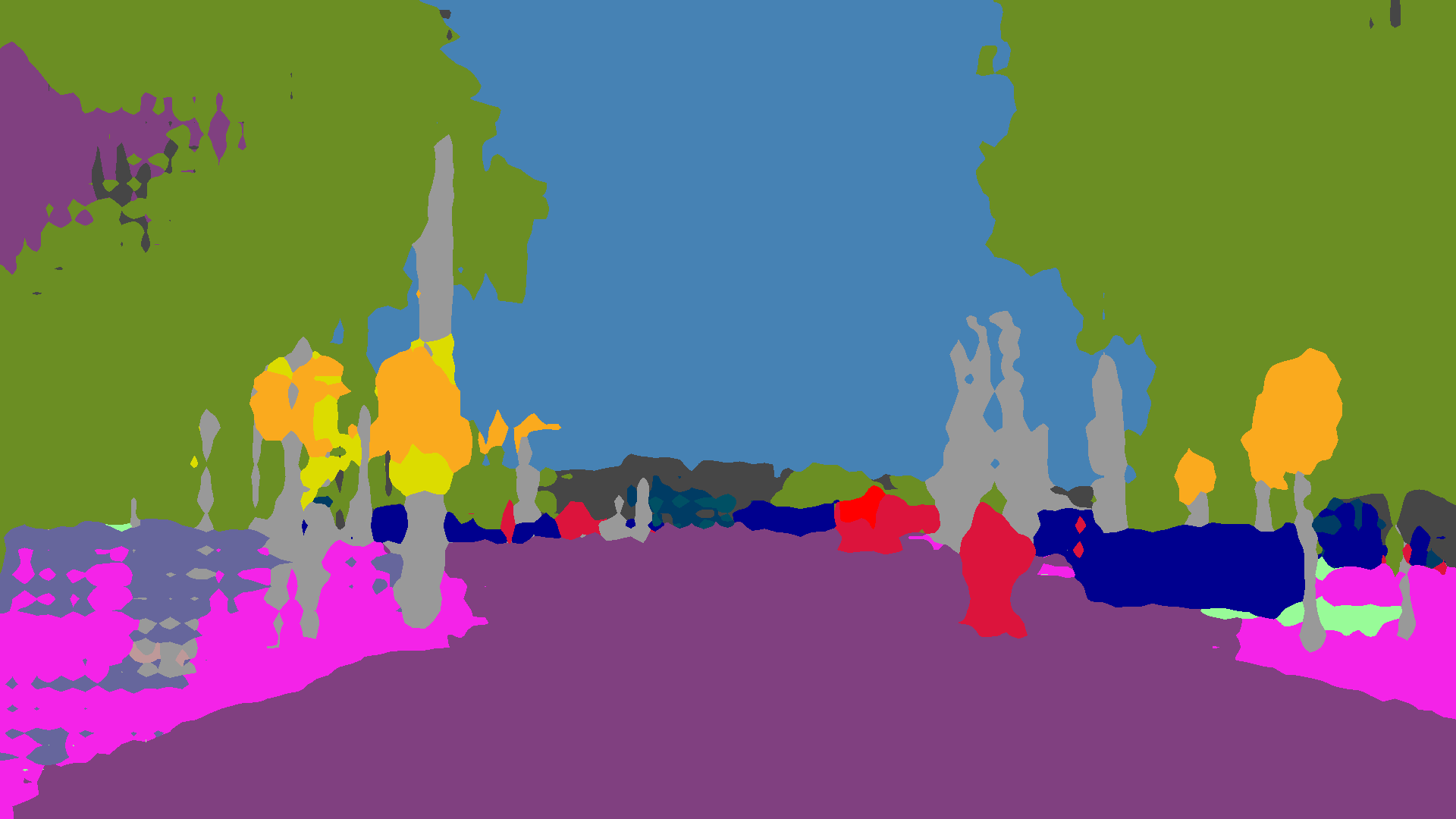} \\
Input & ADVENT & FDA & DACS & DANNet \\
\includegraphics[width=0.2\textwidth, height=1.75cm]{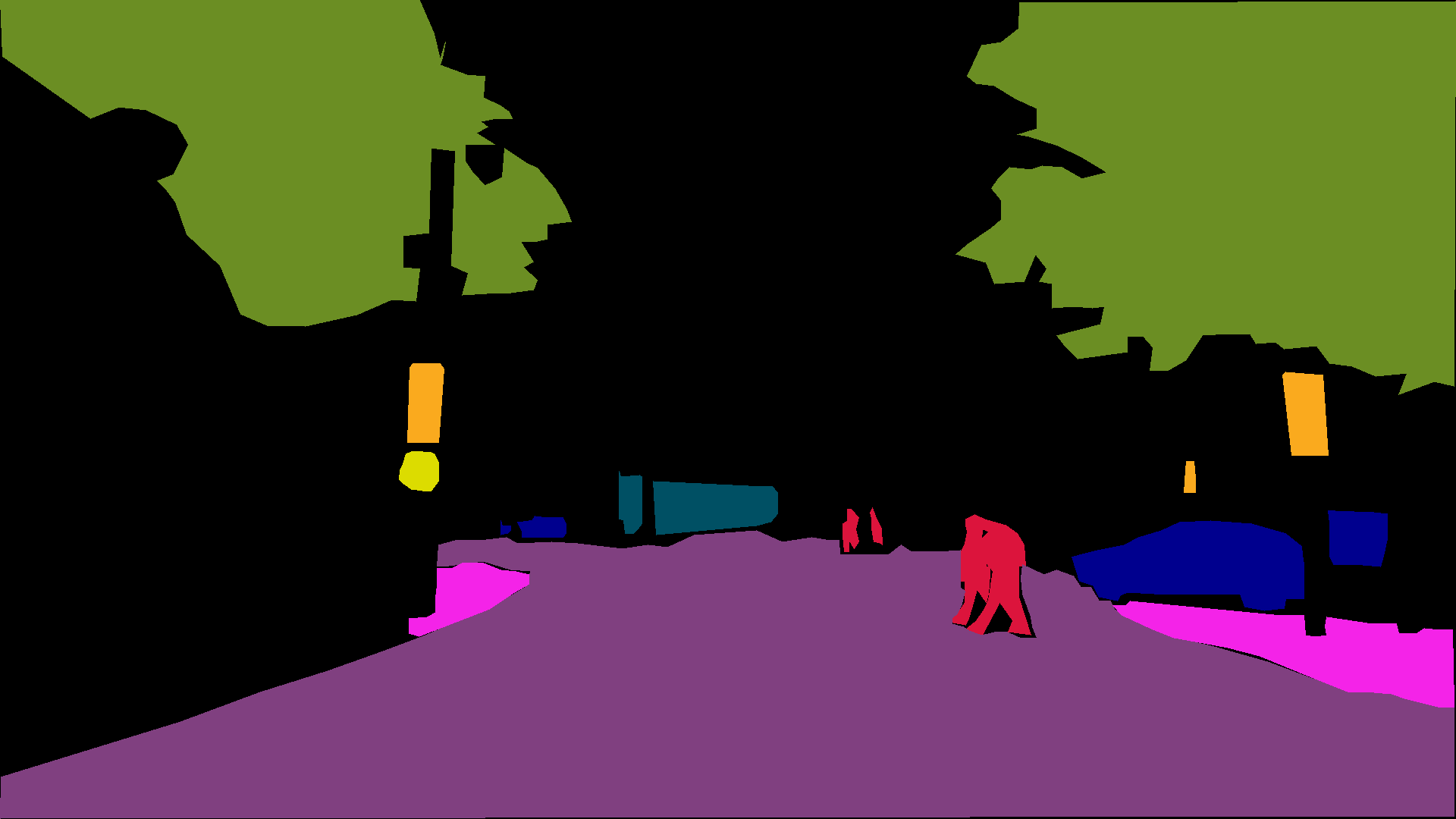} & 
\includegraphics[width=0.2\textwidth, height=1.75cm]{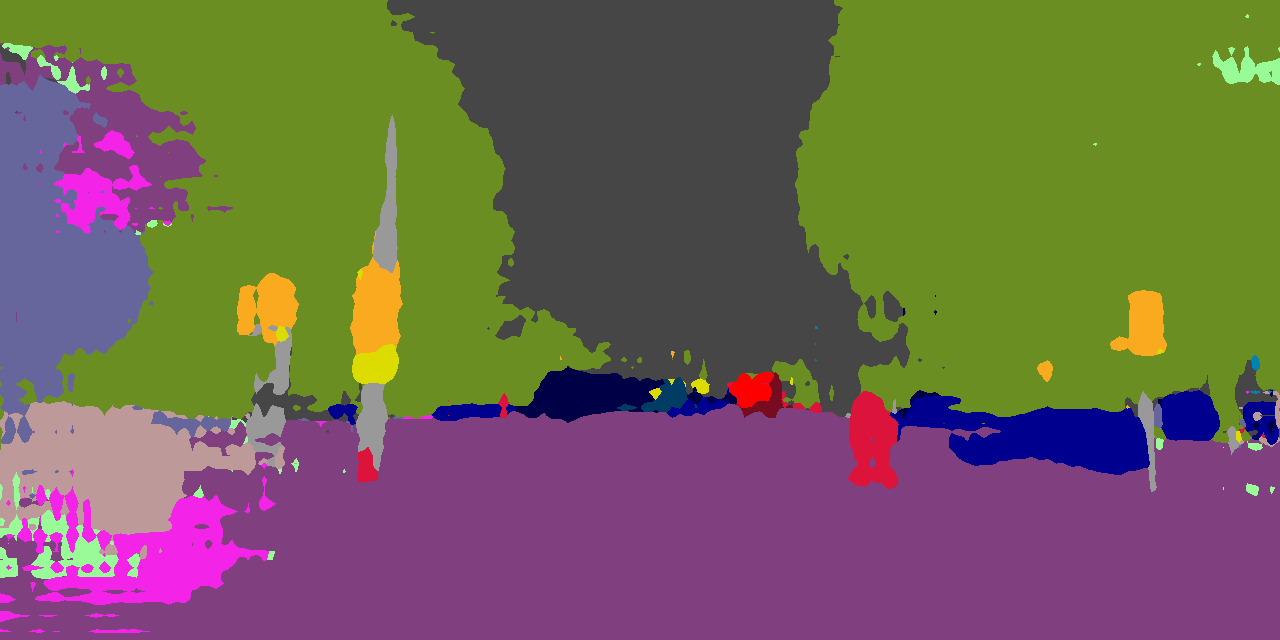} & 
\includegraphics[width=0.2\textwidth, height=1.75cm]{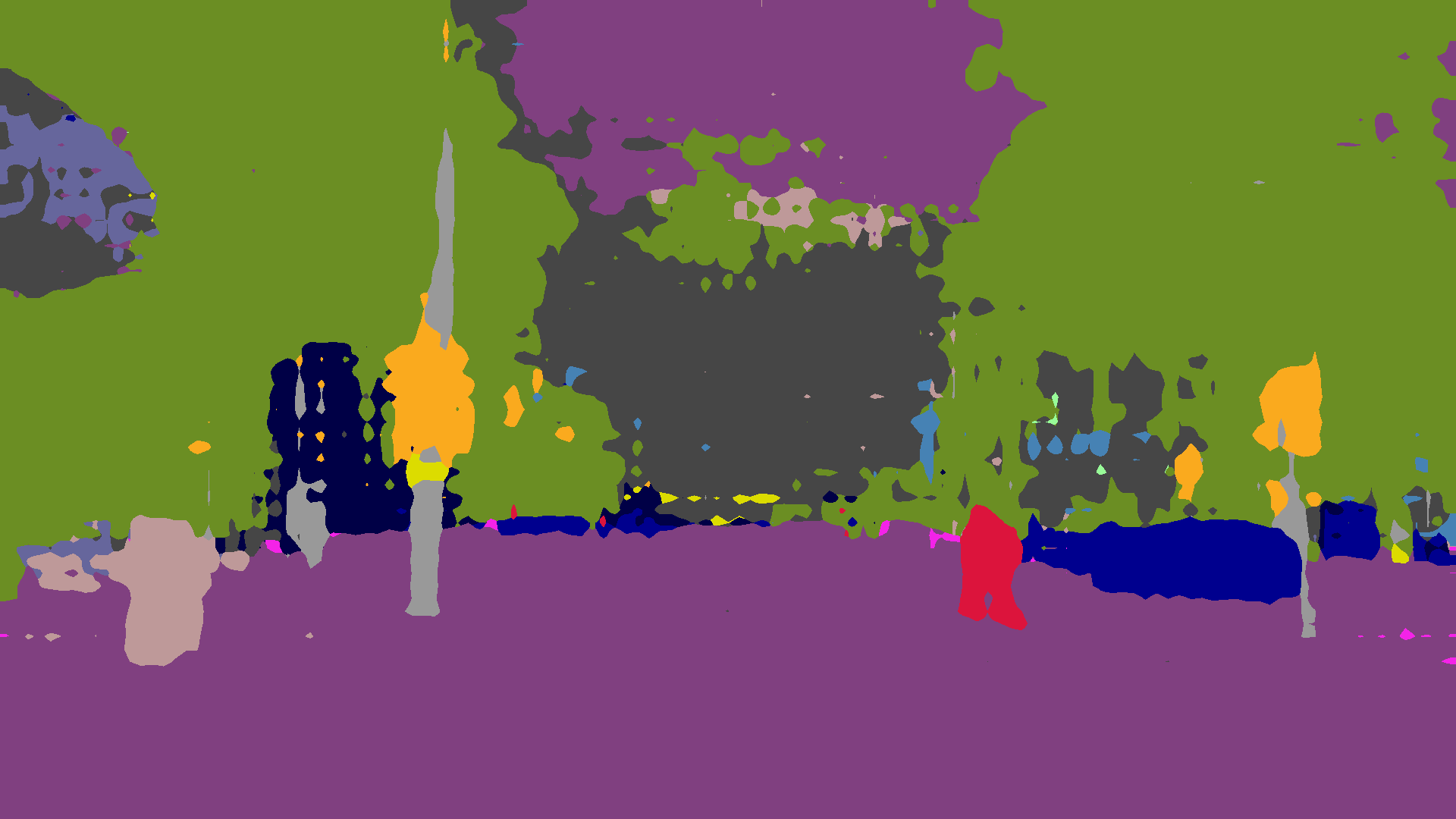} & 
\includegraphics[width=0.2\textwidth, height=1.75cm]{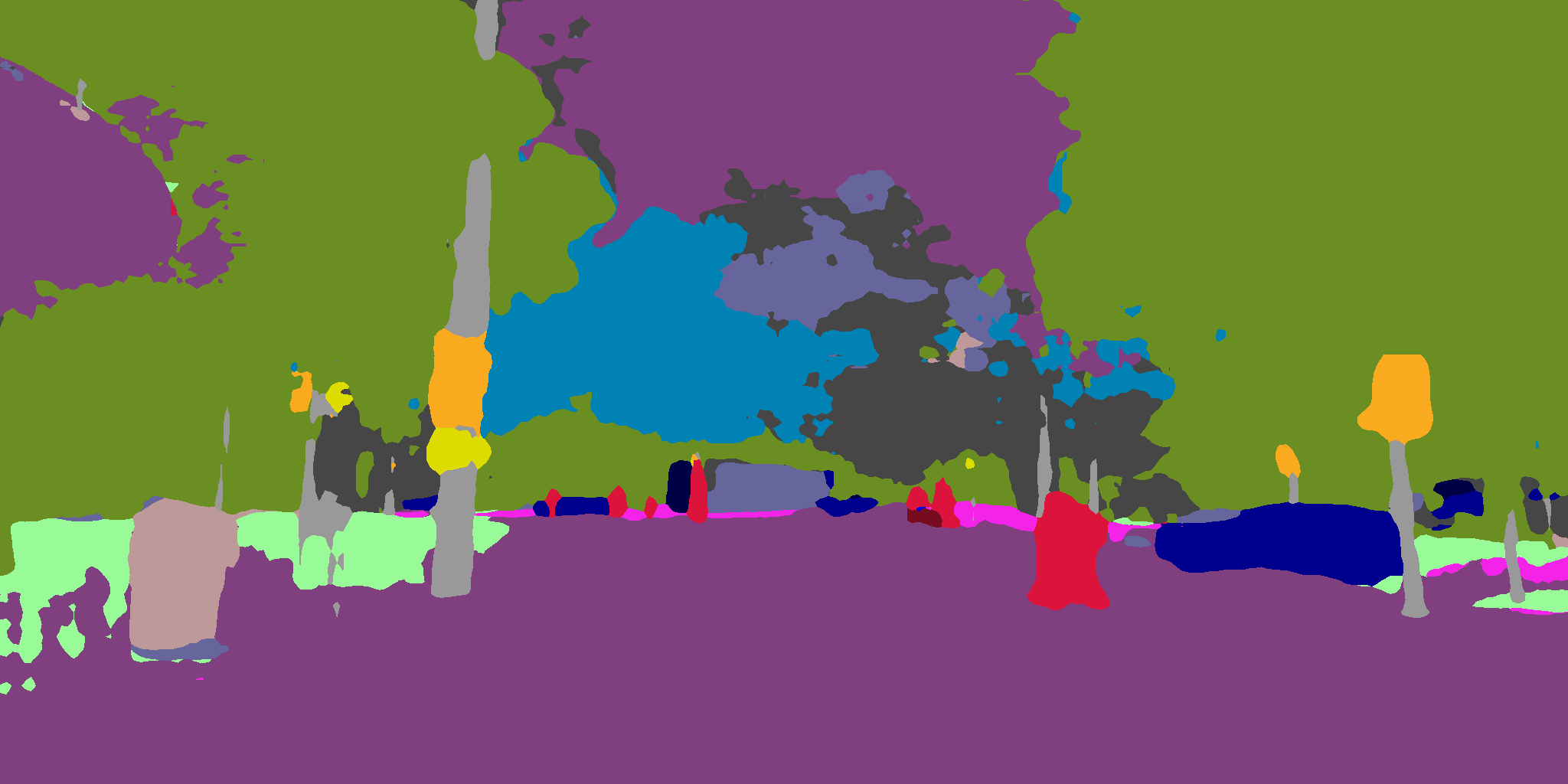} & 
\includegraphics[width=0.2\textwidth, height=1.75cm]{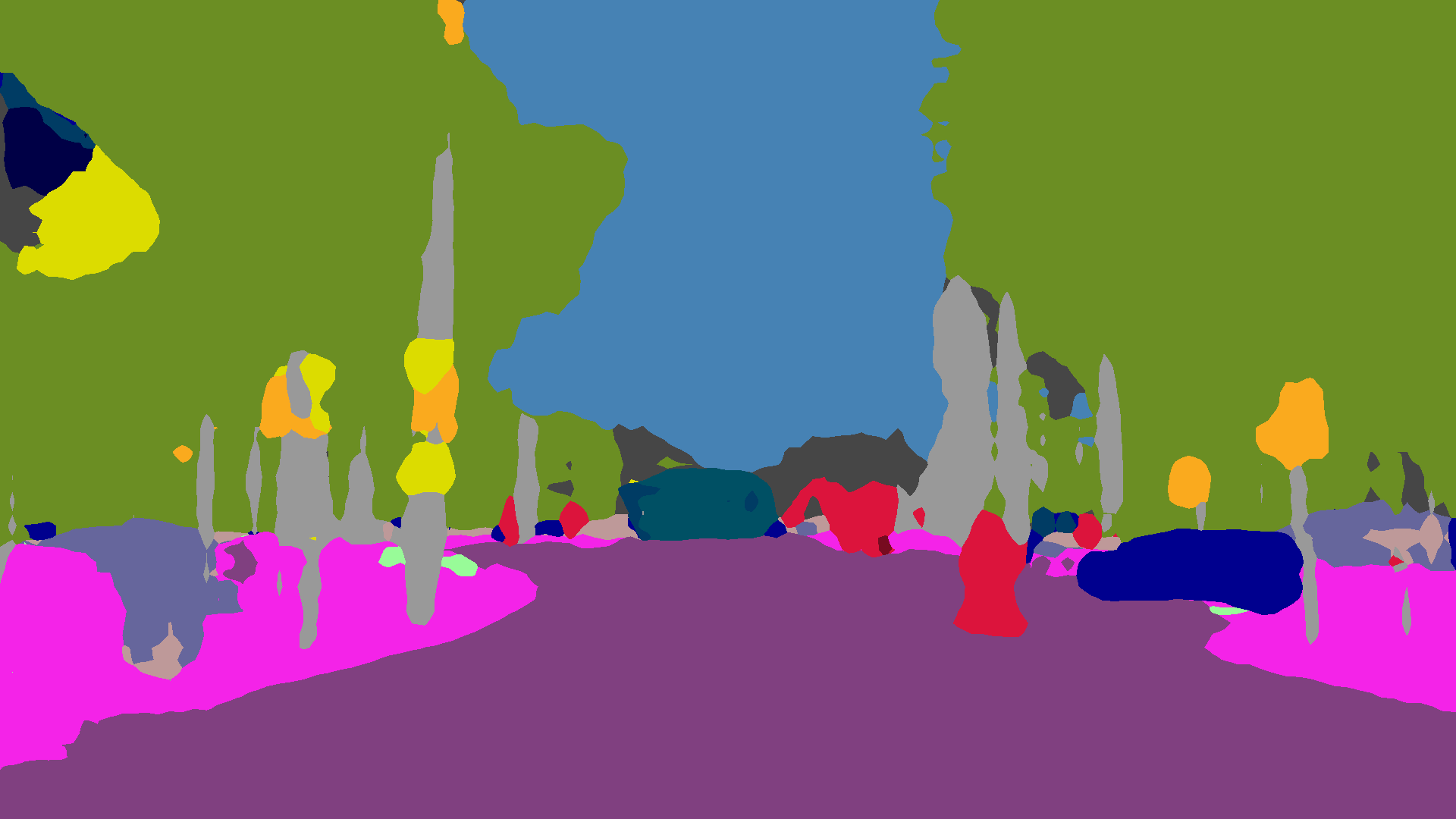} \\
Ground Truth & PixMatch & DISE & ProDA & DGSS \\

\end{tabular}
\end{adjustbox}
\caption{Qualitative Results on Night-Driving Dataset following Cityscapes $\to$ Dark-Zurich scenario.}
\vspace{-4mm}
\label{fig_7}
\end{figure*}

\noindent
\textbf{Unknown Target with low Illumination :} Upon successful demonstration of robustness by proposed framework, we additionally evaluate SoTA algorithms in blind condition where target information is unavailable. We thus use models trained for Cityscapes $\to$ Dark-Zurich scenario and evaluate their performance on Night Driving dataset alongside the proposed framework and summarize the quantitative and qualitative results in Tab. \ref{tab_5} and Fig. \ref{fig_7} respectively. While SoTA algorithms are able to generalize better, their performance is still lower, than observed when following the traditional scenarios of GTAV $\to$ Cityscapes and SYNTHIA $\to$ Cityscapes. We believe if the known target dataset contains diverse illumination conditions, it might alleviate the poor performance observed in low illumination conditions. Nevertheless, such an approach would indeed increase the burden of collecting and diversifying the target domain dataset. From quantitative results we observe DANNet to perform considerably better than its prior, however our mechanism surpasses it by 7.0 mIoU which is observable visually from Fig. \ref{fig_7} wherein the predictions are less noisy compared to PixMatch, DISE, ADVENT, DACS. However from the overall results we can claim that SoTA UDA algorithms are still not able to ensure consistent performance across domains, whereas our proposed mechanism ensures consistent performance without any prior information about target domain simply by using stylized images and contrastive feature alignment.

\noindent
\textbf{Ablation Studies :} 
We first examine the effect of different augmentation techniques and their combination on performance of baseline algorithm that uses both feature representation alignment and iterative style mining. We further examine the effect of replacing iterative style mining with other approaches  specifically (1) random sampled (RS) style, wherein instead of style mining a random style is chosen from the textures and painters dataset to translate the source image into its stylized version and (2) following FDA \cite{yang2020fda}, performing image stylization by simply swapping magnitude or phase components of a randomly sampled stylized image. We follow the same training process as mentioned above and use GTAV dataset for training and Cityscapes dataset for evaluation with performance summarized in Tab. \ref{tab_6}. From the quantitative results we concur that either form of image stylization improves the performance of the baseline algorithm thereby demonstrating the relevance of image stylization in learning domain invariant features. However we observe that stylization performed by ISM results in superior backbone, outperforming RS and FDA demonstrating that simple style transfer by either exchanging frequencies or naively sampling a style, doesnt ensure learning robust domain invariant features. 

\noindent
Apart from the proposed modifications we also examine the effect of individual application of cut-mix and copy-paste augmentations as well as their joint application i.e. style mixing. We observe that similar to observations made for image classification task cut-mix improves the performance of all network by acting as a regularizer and enforcing extraction of efficient representations. Similarly we also witness copy-paste augmentation to improve performance similar in magnitude as cut-mix. However when both these augmentations are applied simultaneously the boost in performance is higher than their individual contribution suggesting these augmentations to be mutually beneficial. 

\noindent
We subsequently examine the effect of using both clustering and cosine similarity loss wherein we observe that when both of these losses are used, they act as complementary to each other boosting the performance of the baseline under all scenarios. Hence demonstrating that clustering and pushing unlike features far apart in latent space improves the performance of the underlying segmentation network. 

\begin{table}[!tb]
\scriptsize
\centering
\renewcommand{\tabcolsep}{3pt} % adjust horizontal space
\begin{adjustbox}{width=0.8\columnwidth}
\begin{tabular}{lcccccc}
\Xhline{3\arrayrulewidth} \noalign{\vskip 1pt}
Cut-Mix & Copy-Paste & $\lambda_1$ & $\lambda_2$ & FDA & RS & Baseline \\
\Xhline{2\arrayrulewidth} \noalign{\vskip 1pt}
& & 1.0 & 0.0 & 39.2 & 41.1 & 42.4 \\
\checkmark & & 1.0 & 0.0 & 40.8 & 47.6 & 48.1 \\
& \checkmark & 1.0 & 0.0 & 41.4 & 48.8 & 48.9 \\
\checkmark & \checkmark & 1.0 & 0.0 & 42.8 & 49.7 & 50.0 \\
\checkmark & \checkmark & 1.0 & 1.0 & 46.9 & 49.9 & 53.6 \\
\Xhline{3\arrayrulewidth} \noalign{\vskip 1pt}
\end{tabular}
\end{adjustbox}
\caption{Ablation Studies for Different Mechanisms}
\vspace{-5mm}
\label{tab_6}
\end{table}

% We further examine the benefits of performing iterative style mining and summarize the performance landscape of varying number of styles within style bank from either paintings or textures in Fig. \ref{fig_7}. From the landscape we can infer that after a certain limit (11 for paintings and 8 for materials) the effect of varying the number of styles doesn't improve the performance of the network. Nevertheless allowing us to achieve domain generalization properties by ensuring extraction of domain invariant features. 

% \begin{figure}[!th]
% \includegraphics[width=\columnwidth, height=4.0cm]{Images/.jpg}
% \caption{Performance of SS model on GTAV$\to$Cityscapes when using different style configurations}
% \label{fig_7}
% \vspace{-4mm}
% \end{figure}

\vspace{-2mm}
\section{Conclusion}
This paper demonstrates that domain generalization for semantic segmentation could be achieved without any information about the target domain by simply identifying styles and textures that maximize the domain gap w.r.t. the source. Apart from generating diverse training distribution, we also perform latent representation alignment wherein categorical features are clustered together irrespective of the domain gap. Doing so ensures that the semantic segmentation algorithm learns structural details for making robust predictions. To examine the domain generalization performance, we conducted quantitative and qualitative experiments with SoTA UDA algorithms and achieved comparable performance to SoTA, however, the performance consistency was observed even for scenarios where SoTA UDA performed poorly, such as low illumination conditions.  

\bibliographystyle{aaai22}
\bibliography{egbib.bib}
\end{document}